\documentclass[10pt,twocolumn,letterpaper]{article}
\usepackage[pagenumbers]{cvpr}
\usepackage[T1]{fontenc}
\usepackage[utf8]{inputenc}
\usepackage{mathtools}
\usepackage{array}
\usepackage{tabularx}
\usepackage{multirow}
\usepackage{colortbl}
\usepackage{microtype}

\newcommand{\E}{\mathbb{E}}

\newcommand{\D}{\mathcal{D}}
\newcommand{\A}{\mathcal{A}}

\definecolor{cvprblue}{rgb}{0.21,0.49,0.74}
\usepackage[breaklinks,colorlinks,allcolors=cvprblue]{hyperref}
\hypersetup{pdftitle={What Makes a Good Move? Learning to Act from Experience in Visual Worlds},
  pdfauthor={Cehao Yang, Ziyi He, Jing Yi, Jian Guo, Hui Xiong}}
\title{What Makes a Good Move? Learning to Act from Experience in Visual Worlds}
\author{Cehao Yang\thanks{Work done during internship at \mbox{LIGHTSPEED}.}\\
HKUST(GZ)\\
{\small\ttfamily cyang289@connect.hkust-gz.edu.cn}
\and
Ziyi He\\
HKU\\
{\small\ttfamily max1223@connect.hku.hk}
\and
Jing Yi\thanks{Corresponding author.}\\
LIGHTSPEED\\
{\small\ttfamily yijing-v@whu.edu.cn}
\end{tabular}\\[.5em]
\begin{tabular}[t]{c}
Jian Guo\\
IDEA Research\\
{\small\ttfamily guojian@idea.edu.cn}
\and
Hui Xiong\\
HKUST(GZ)\\
{\small\ttfamily xionghui@ust.hk}
}
\makeatletter
\patchcmd{\@maketitle}{\vspace*{24pt}}{\vspace*{10pt}}{}{%
  \PackageError{lightact}{Unable to adjust the title-to-author spacing}{}%
}
\apptocmd{\@maketitle}{\vspace{-14pt}
\begin{minipage}{\textwidth}
\centering
\includegraphics[width=0.93\linewidth,trim=0 0 0 26bp,clip]{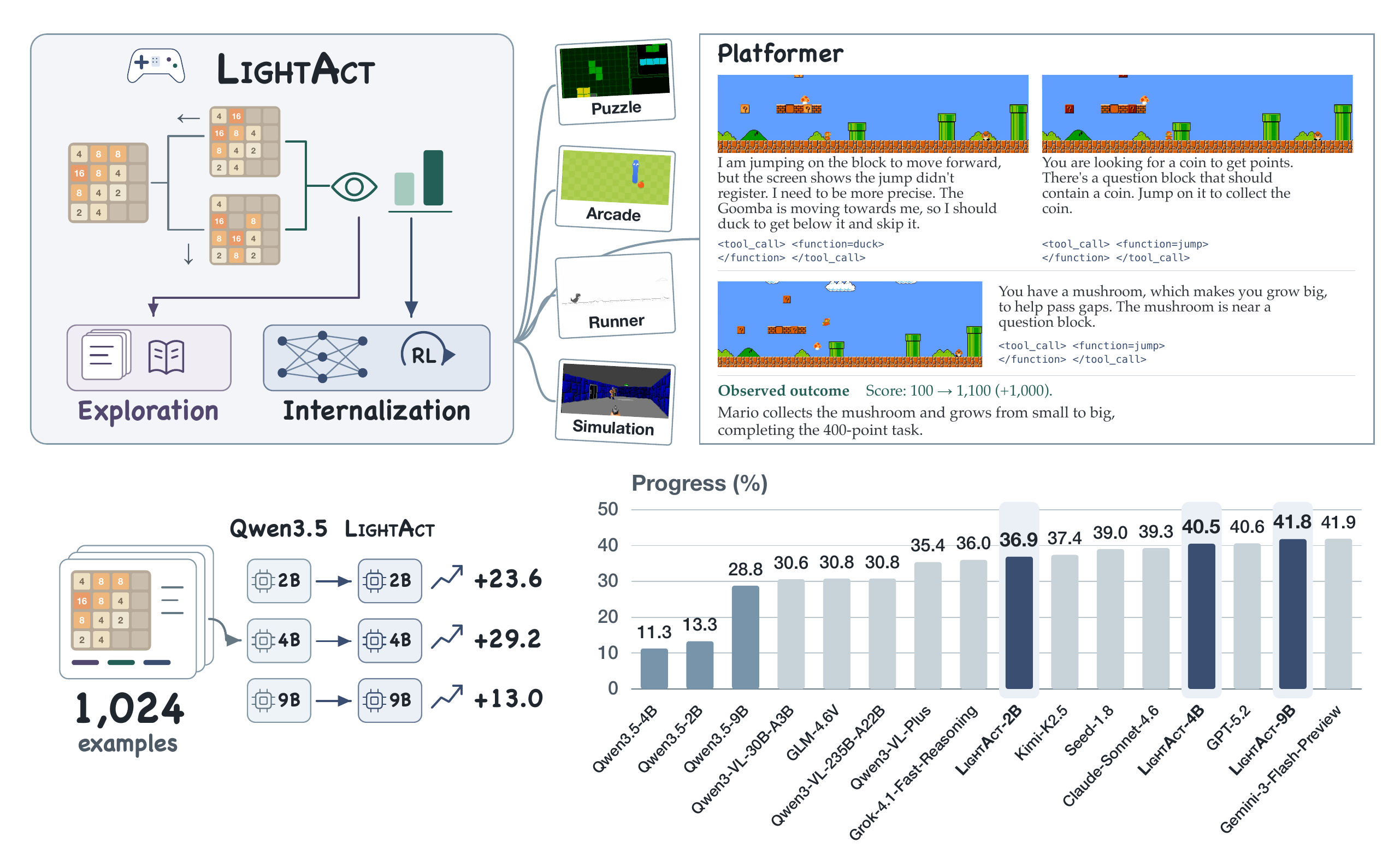}
\captionof{figure}{Learning from source experience with \textsc{LightAct}.}
\label{fig:teaser}
\end{minipage}
\par\vspace{7pt}
}{}{%
  \PackageError{lightact}{Unable to add the first-page teaser}{}%
}
\makeatother
\begin{document}
\maketitle
\begin{abstract}
Learning visual policies requires turning interaction experience into guidance for individual decisions. Trajectory outcomes obscure action quality, while unguided exploration can revisit familiar situations. We present \textsc{LightAct}, an experience-learning framework that connects skill-guided exploration with policy internalization. Interaction segments are summarized into experience entries and consolidated into a game-specific skill that guides root collection toward broader state coverage. At each root, candidate actions are compared through short rollouts under a shared frozen policy. Environment-reported score gains are averaged across continuations and normalized into relative action rewards, producing reusable supervision for reinforcement learning. Policy training uses a compact dataset of just 1,024 decision roots collected in a single source game and paired with their action rewards. Across 170 GameWorld tasks spanning 34 games, \textsc{LightAct} improves overall progress by 23.6, 29.2, and 13.0 percentage points at the 2B, 4B, and 9B scales, respectively, with our strongest model competitive with reported proprietary-model results. Ablations show broader root-state coverage with skill guidance and consistent gains across three policy optimizers. Case studies show how rollouts distinguish actions with identical immediate gains.
\end{abstract}

\section{Introduction}
Vision-language models (VLMs) increasingly serve as agents that connect visual perception and language reasoning to action in interactive environments~\cite{lma_survey,agentic_mllm_survey,cradle,optimus2}. Video games expose the demands of this process: an agent must recognize the current situation, choose an executable move, and adapt as its actions change the environment~\cite{balrog,gameworld,hu_game_agents}. Interaction also provides experience from which to improve. Prior work retains such experience as contextual guidance~\cite{jarvis1,ical} or uses it to update policy parameters through reinforcement learning~\cite{digirl,lmgame}.

The central challenge is to turn this experience into guidance about individual decisions. A trajectory records the actions taken and their realized outcomes, leaving alternatives at the same decision point unobserved~\cite{logged_bandit_feedback}. Its final outcome combines many choices: a useful move can be followed by a mistake, while a poor move can be rescued later. Environment scores summarize these outcomes, but do not directly reveal which action would have been preferable at a particular state~\cite{temporal_credit_assignment}. The states encountered during exploration introduce a second limitation. Repeatedly visiting familiar situations concentrates experience on a narrow range of decisions~\cite{go_explore,intelligent_go_explore}.

These observations suggest organizing experience around both the states reached and the choices available there. Summaries of prior interaction can guide exploration toward a broader set of decision states. At a selected state, executing alternative actions under a shared continuation policy makes their subsequent outcomes comparable. Averaging score gains over short rollouts then yields an estimate of each action's value, which can be converted into supervision for policy learning.

We present \textsc{LightAct}, an experience-learning framework that combines \emph{skill-guided exploration} with \emph{policy internalization}. We partition preliminary trajectories collected by a frozen pretrained VLM into short interaction segments. The model summarizes each segment into an experience entry describing the observed conditions, behavior, and outcome, then consolidates these entries into a game-specific textual skill. The skill guides subsequent exploration by providing reusable advice on effective behavior and recurring failures. We retain distinct visited states and their visual decision contexts as a bank of decision roots.

At each root, we compare the available actions through repeated short continuations under the same frozen policy. The resulting environment-reported score gains are averaged and normalized within the root to produce relative action rewards. Pairing these rewards with the collected decision contexts forms a fixed training set. Reinforcement learning then updates the policy by sampling responses to these contexts and assigning rewards to their parsed actions.

\begin{figure}[t]
\centering
\includegraphics[width=\columnwidth]{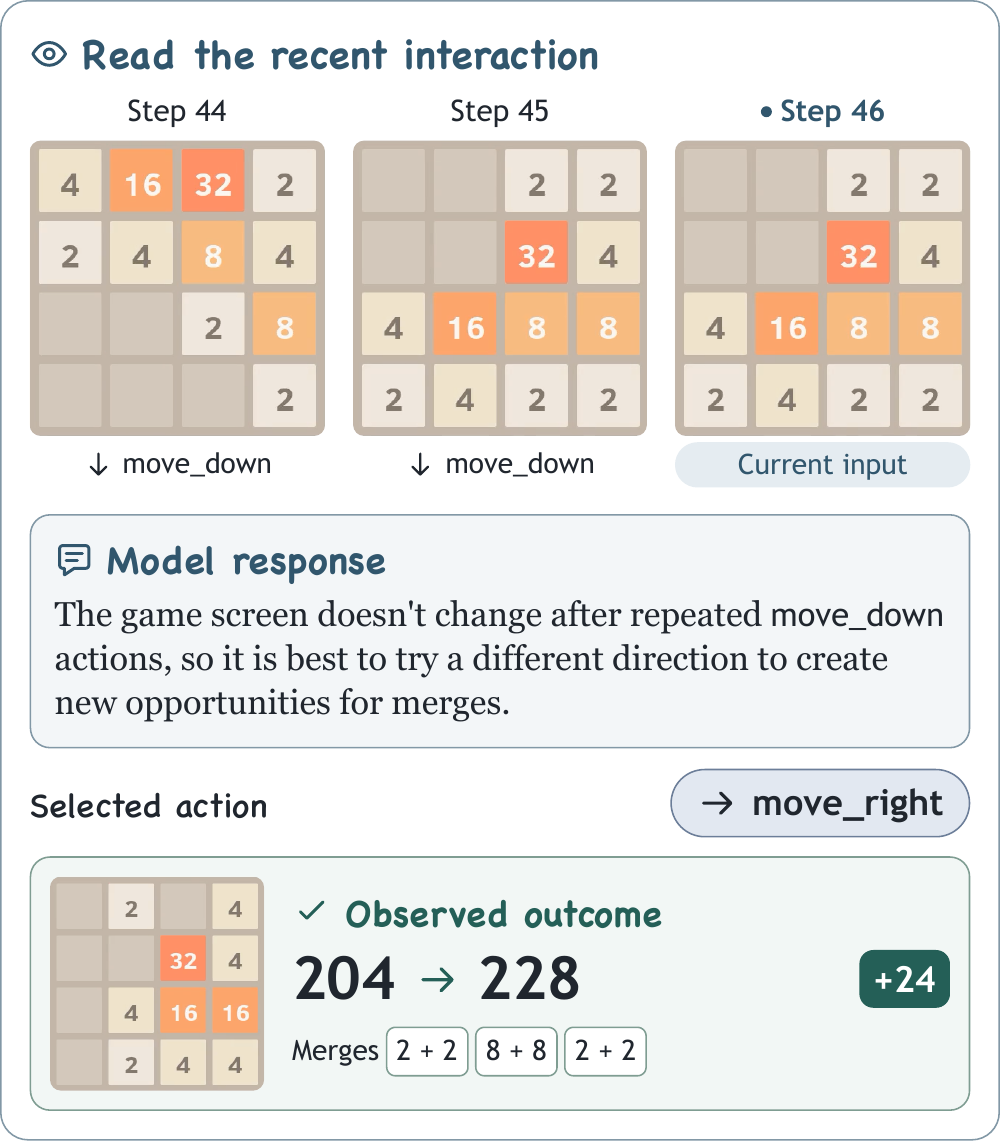}
\caption{\textbf{A 2048 decision by \textsc{LightAct}-2B.} When repeated downward moves no longer change the board, the policy switches to moving right and gains 24 points.}
\label{fig:intro_decision}
\end{figure}

We instantiate \textsc{LightAct} at the 2B, 4B, and 9B scales using Qwen3.5~\cite{qwen35} and evaluate on all 170 GameWorld tasks across 34 games and five categories~\cite{gameworld}. With experience collected in 2048, overall progress improves by 23.6, 29.2, and 13.0 percentage points at the three scales, respectively. \textsc{LightAct}-2B reaches 36.9\% progress and 13.9\% success, exceeding the pretrained 9B baseline on both metrics, while \textsc{LightAct}-9B achieves 41.8\% progress, competitive with reported proprietary-model results. Further experiments show broader root-state coverage with skill guidance, gains across three RL optimizers, and benefits from source-game experience across game categories. Qualitative examples illustrate how rollout comparisons reveal action preferences that immediate gains alone cannot distinguish.

Our contributions are threefold:
\begin{itemize}
    \item We introduce an experience-learning framework that connects skill-guided source exploration with reinforcement learning from action-level feedback.
    \item We construct reusable action-reward tables by comparing continuations from shared decision roots under a frozen policy and normalizing their environment-reported score gains.
    \item We demonstrate improvements across three model scales on GameWorld and examine how skill guidance, source environments, and policy optimizers contribute to learning from interaction.
\end{itemize}

\section{Related Work}
\label{sec:related}
\subsection{Vision-Language Game Agents}
Video games test whether visual understanding supports sequential decisions~\cite{vmage}. Cradle~\cite{cradle} combines perception, planning, and memory to operate through screenshots and keyboard or mouse controls. BALROG~\cite{balrog} evaluates language and vision-language agents across games and reveals weaknesses in visually grounded interaction. VideoGameBench~\cite{videogamebench} examines play from raw visual input under real-time and paused interaction, separating decision quality from the effects of inference latency. GameWorld~\cite{gameworld} provides browser games with standardized interfaces and state-verifiable task evaluation. Complementing these benchmarks, lmgame-Bench~\cite{lmgame} studies perception and memory scaffolds and reports that reinforcement learning on individual games can transfer beyond the training environment. Our work examines how source-game interaction supports such policy learning.

\subsection{Experience Learning of VLMs}
VLM agents learn from accumulated interaction through contextual guidance and parameter updates. External memory retains experience for retrieval at decision time~\cite{echo_minecraft}, while parameter updates encode it in the policy~\cite{peam}. JARVIS-1~\cite{jarvis1} retrieves multimodal Minecraft experience to support planning, while ELITE~\cite{elite} extracts reusable strategies from trajectories and retrieves them according to task intent. ICAL~\cite{ical} refines suboptimal demonstrations through execution and human feedback, annotating causal relations, state changes, and subgoals; the resulting examples support both retrieval and supervised fine-tuning. For reinforcement learning, DigiRL~\cite{digirl} adapts a pretrained VLM for device control through offline initialization and online interaction, using a VLM evaluator for feedback. VAGEN~\cite{vagen} improves explicit state estimation and transition prediction in multi-turn visual interaction through reinforcement learning. We connect experience-guided source exploration with policy internalization.

\subsection{Rollout-Based Policy Learning}
Rollouts estimate the value of an action from the outcomes that follow it. Monte Carlo policy improvement compares candidate moves by continuing each with a common base policy~\cite{tesauro1996}. In combinatorial optimization and stochastic scheduling, heuristic continuations estimate the future cost of each choice~\cite{bertsekas1997,bertsekas1999}. Sparse sampling~\cite{kearns1999} estimates action values over sampled lookahead trees, while UCT~\cite{kocsis2006} uses a bandit rule to allocate simulations among branches. Rollout comparisons also provide supervision for policy learning~\cite{ipr1}. Classification-based policy iteration~\cite{lagoudakis2003} uses the preferred action at a sampled state as a training label. AggreVaTe~\cite{ross2014} trains on expert cost-to-go estimates, while LOLS~\cite{chang2015} evaluates decisions under both reference and learned policies. For language models, LATS~\cite{lats} guides tree search with model evaluations and environment feedback, and Math-Shepherd~\cite{mathshepherd} uses sampled completions to assess intermediate reasoning steps. We use score gains from a shared game state to supervise action selection.

\section{Method}
\label{sec:method}
\begin{figure*}[t]
    \centering
    \includegraphics[width=\textwidth]{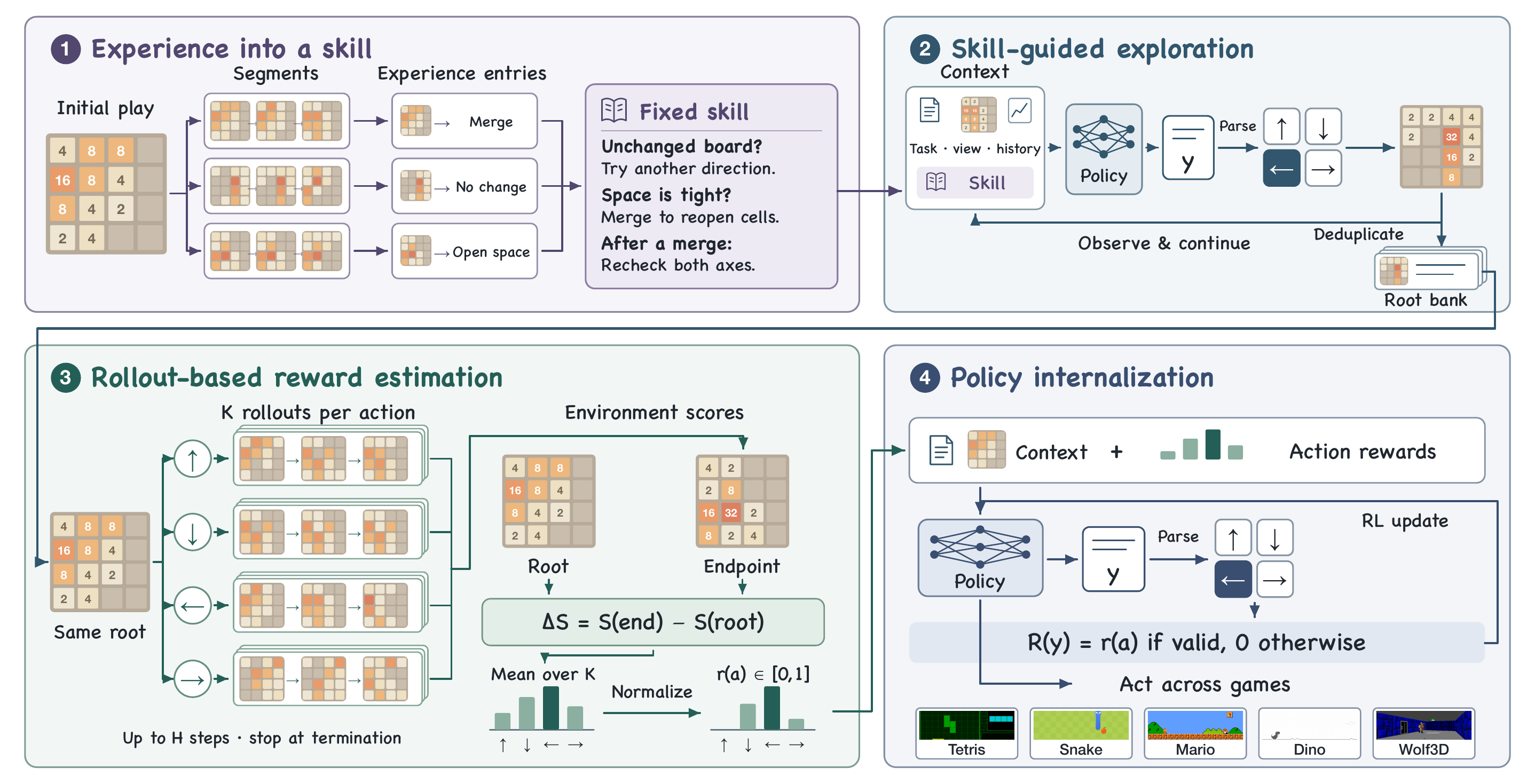}
    \caption{Overview of \textsc{LightAct}. Interaction segments are summarized into experience entries, then consolidated into a source-game skill that guides root-state collection. At each root, short rollouts compare candidate actions through their subsequent score gains. These gains are normalized into action-level rewards for reinforcement learning, internalizing the collected experience in a policy evaluated across games.}
    \label{fig:method_overview}
\end{figure*}

Experience gathered during interaction can guide subsequent decisions through both contextual guidance and policy learning. Building on this view, \textsc{LightAct} (Fig.~\ref{fig:method_overview}) distills source-game trajectories into textual skills for exploration and compares action outcomes at the collected roots to construct action-level rewards for policy internalization.

\subsection{Decision-Making in Game Environment}
In GameWorld~\cite{gameworld}, an agent pursues a game-specific objective through sequential interaction. At each decision, it receives the game rules, task instruction, role information, and available semantic actions, together with the current screenshot and recent observation--action history. The agent generates a response, parses it into an executable action, observes its effect on the game, and continues toward the objective. This interaction links visual understanding to decisions whose consequences unfold over subsequent steps.

For a game environment $e$ with a fixed, finite action space $\A_e$, let $s_t\in\mathcal{S}_e$ be its state, $a_t\in\A_e$ an action, and $o_t=\Omega_e(s_t)$ the visual observation. Interaction follows
\begin{equation}
 s_{t+1}\sim p_e(\cdot\mid s_t,a_t),\qquad
 x_t=(g,o_t,h_t),
\end{equation}
where $g$ denotes the task specification together with the game instructions, $o_t$ is the current screenshot, and $h_t$ contains the recent observations and actions. The instructions describe the objective, rules, acting role, and semantic action interface. A task-specific evaluator $\mathrm{Eval}_{e,g}$ determines whether the current environment state satisfies the task's completion criteria, with $\mathrm{Eval}_{e,g}(s_t)=1$ indicating success. Let $\pi_\theta$ denote the policy parameterized by $\theta$. Its pretrained parameters $\theta_0$ define the frozen initial policy $\pi_0=\pi_{\theta_0}$. The policy generates a response $y_t\sim\pi_\theta(\cdot\mid x_t)$ from this context. The parser $\mathcal P$ attempts to extract an executable action from $y_t$: successful parsing yields $a_t=\mathcal P(y_t)$, while invalid responses yield no executable action. For games with multiple acting roles, the other roles' behavior is part of the transition process.

\subsection{Exploration and Root-State Collection}
\label{sec:rootcollection}
The states reached during source interaction determine the decisions available for policy learning. We collect trajectories with a collection policy $\beta_e$, sampling responses $y_t\sim\beta_e(\cdot\mid x_t)$ and executing successfully parsed actions. The basic collection policy is $\beta_e(\cdot\mid x_t)=\pi_0(\cdot\mid x_t)$.

Among the nonterminal states visited during collection, we retain the first occurrence under an environment-specific equivalence criterion. For temporally ordered candidates indexed by $\mathcal C$, the selected indices are
\begin{equation}
 \mathcal I=\{i\in\mathcal C:
 \operatorname{Deduplicate}_e(s_i)=1\}.
 \label{eq:root_selection}
\end{equation}
Each selected state is stored with its decision context $x_i$. The selected states form the root bank:
\begin{equation}
 \D_{\mathrm{root}}=\{(s_i,x_i):i\in\mathcal I\},
 \qquad N=|\mathcal I|.
\end{equation}
Their original contexts $x_i=(g,o_i,h_i)$ and the action rewards constructed below form the training examples for policy internalization.

\subsection{Skill-Guided Exploration}
\label{sec:experience}
\label{sec:exploration}
To guide exploration toward a richer set of decision states, we first collect preliminary trajectories $\mathcal T_e^{\mathrm{init}}$ with $\pi_0$ and partition them into short \emph{interaction segments} $\{\sigma_j\}_{j=1}^{M}$. Each segment contains consecutive visual observations, executed actions, and their resulting changes. Under an entry-summarization prompt, the frozen model distills each segment into an \emph{experience entry}:
\begin{equation}
 \eta_j=F_{\theta_0}^{\mathrm{entry}}(g,\sigma_j).
\end{equation}
Each entry captures the observed conditions, behavior, and outcome within its segment.

The same model then consolidates the entries across segments into one textual skill for environment $e$:
\begin{equation}
 \kappa_e=F_{\theta_0}^{\mathrm{skill}}(g,\{\eta_j\}_{j=1}^{M}).
 \label{eq:skill_extraction}
\end{equation}
Through a skill-consolidation prompt, recurring observations are combined into concise guidance on useful behavior and failure patterns, tied to the conditions under which they arise. This guidance helps the agent reuse effective interactions and adjust its behavior when exploration stalls.

We include this skill in the collection policy's decision context:
\begin{equation}
 y_t\sim\beta_e(\cdot\mid x_t)
 =\pi_0(\cdot\mid x_t,\kappa_e).
\end{equation}
The policy interprets the guidance using the current observation and recent history. The skill and model parameters remain fixed throughout guided collection.

We construct the root bank $\D_{\mathrm{root}}$ from the guided trajectories using the root-selection procedure above.

\subsection{Rollout-Based Action Value Estimation}
\label{sec:branch}
For each decision root, we restore the same environment state $s_i$ and decision context $x_i$ before evaluating each candidate action. Starting from $s_i$, we execute action $a$ and let $\pi_0$ select the subsequent actions. The resulting branch contains at most $H$ decisions, including the initial action, and ends at the first episode termination if it occurs earlier:
\begin{equation}
 B_{i,a}\sim\mathcal B_H(\cdot\mid s_i,x_i,a,\pi_0).
\end{equation}
At each subsequent decision, $\pi_0$ conditions on the task, current observation, and recent interaction history within the branch. Using the same frozen policy across actions and roots provides a shared basis for comparing their outcomes.

Let $S_e(s)$ denote the game score returned by the environment at state $s$ and $s_B$ the final state of branch $B$. A continuation from $s_i$ contributes the score gain
\begin{equation}
 G_e(s_i,B)=S_e(s_B)-S_e(s_i).
 \label{eq:scoregain}
\end{equation}
For a terminal branch, $s_B$ is the state before reset. The finite-horizon action value and its estimate from $K$ continuations are
\begin{align}
 Q_H^{\pi_0}(i,a)&=\E_{B_{i,a}}[G_e(s_i,B_{i,a})],\\
 \widehat Q_i(a)&=\frac{1}{K}\sum_{k=1}^{K}
 G_e(s_i,B_{i,a}^{(k)}).
 \label{eq:qestimate}
\end{align}
The estimate measures score gain after choosing $a$ and continuing under $\pi_0$ for the specified horizon.

\subsection{Reward Shaping}
\label{sec:rewardshaping}
The estimated score gains $\widehat Q_i(a)$ quantify the outcomes of candidate actions in the game's score units. To express action preferences on a common reward scale, we normalize these values within each root:
\begin{equation}
 r_i(a)=
 \begin{cases}
 \dfrac{\widehat Q_i(a)-q_i^-}{q_i^+-q_i^-}, & q_i^+>q_i^-,\\[4pt]
 0, & q_i^+=q_i^-,
 \end{cases}
 \label{eq:normalization}
\end{equation}
where $q_i^-=\min_{a\in\A_e}\widehat Q_i(a)$ and $q_i^+=\max_{a\in\A_e}\widehat Q_i(a)$. This normalization preserves the estimated ordering of actions and expresses their relative quality within a shared decision context. When all action values coincide, the root supplies no preference signal and all action rewards are zero.

\setcounter{dbltopnumber}{1}
\begingroup
\definecolor{bggray}{HTML}{F2F2F2}
\definecolor{textgray}{HTML}{4A4A4A}
\newcommand{\modellogo}[2]{\raisebox{-0.05em}{\includegraphics[height=0.9em]{figures/logos/#1}} #2}
\begin{table*}[t]
\scriptsize
\centering
\caption{\textbf{Main results on GameWorld.} SR and PG (\%) are reported by category and overall. All three \textsc{LightAct} models are trained with GRPO using experience collected in 2048. Qwen3.5 and \textsc{LightAct} results are averaged over three evaluation runs per task. Published baseline results are taken from GameWorld~\cite{gameworld}.}
\setlength{\tabcolsep}{0.25mm}
\renewcommand{\arraystretch}{1.05}

\newcolumntype{S}{>{\centering\arraybackslash}p{0.61cm}}
\newcolumntype{M}{>{\centering\arraybackslash}p{0.75cm}}

{
\resizebox{\linewidth}{!}{
\begin{tabular}{l *{2}{S} *{2}{M} *{6}{S} | *{2}{S}}
\toprule
\multirow{2}{*}{\textbf{Model}} &
\multicolumn{2}{c}{\textbf{Arcade}} &
\multicolumn{2}{c}{\textbf{Platformer}} &
\multicolumn{2}{c}{\textbf{Puzzle}}  &
\multicolumn{2}{c}{\textbf{Runner}} &
\multicolumn{2}{c|}{\textbf{Simulation}} &
\multicolumn{2}{c}{\textbf{Overall}} \\
\cmidrule(lr){2-3} \cmidrule(lr){4-5} \cmidrule(lr){6-7} \cmidrule(lr){8-9} \cmidrule(lr){10-11} \cmidrule(lr){12-13}

 & $\mathcal{SR}$ & $\mathcal{PG}$ & $\mathcal{SR}$ & $\mathcal{PG}$ & $\mathcal{SR}$ & $\mathcal{PG}$ & $\mathcal{SR}$ & $\mathcal{PG}$ & $\mathcal{SR}$ & $\mathcal{PG}$ & $\mathcal{SR}$ & $\mathcal{PG}$ \\

\midrule
\multicolumn{13}{l}{\cellcolor{bggray}\textcolor{textgray}{\textbf{\textit{Human}}}}\\
Novice Player  & 45.7 & 55.5 & 60.0 & 65.6 & 51.4 & 63.1 & 60.0 & 72.0 & 60.0 & 62.0 & 55.3 & 64.1 \\
Expert Player  & 65.7 & 73.9 & 85.0 & 88.0 & 68.6 & 77.1 & 82.5 & 87.8 & 85.0 & 86.0 & 77.1 & 82.6 \\

\midrule
\multicolumn{13}{l}{\cellcolor{bggray}\textcolor{textgray}{\textbf{\textit{Generalist Multimodal Agents}}}}\\
\modellogo{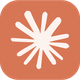}{Claude-Sonnet-4.6} & 5.7 & 28.3 & 22.5 & 37.0 & 25.7 & 51.5 & 30.0 & 51.9 & 15.0 & 16.6 & 20.6 & 39.3 \\
\modellogo{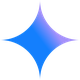}{Gemini-3-Flash-Preview} & 5.7 & 26.3 & 25.0 & 41.2 & 25.7 & 54.8 & 32.5 & 55.4 & 10.0 & 21.1 & 21.2 & 41.9 \\
\modellogo{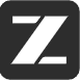}{GLM-4.6V} & 8.6 & 22.8 & 20.0 & 33.9 & 5.7 & 29.1 & 27.5 & 49.1 & 0.0 & 5.3 & 14.1 & 30.8 \\
\modellogo{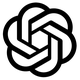}{GPT-5.2} & 8.6 & 29.3 & 22.5 & 36.7 & 28.6 & 56.2 & 27.5 & 52.6 & 10.0 & 16.9 & 20.6 & 40.6 \\
\modellogo{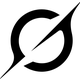}{Grok-4.1-Fast-Reasoning} & 8.6 & 23.7 & 22.5 & 37.3 & 14.3 & 46.6 & 25.0 & 49.0 & 5.0 & 10.4 & 16.5 & 36.0 \\
\modellogo{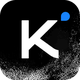}{Kimi-K2.5} & 8.6 & 26.4 & 20.0 & 35.3 & 25.7 & 51.4 & 27.5 & 49.7 & 5.0 & 11.7 & 18.8 & 37.4 \\
\modellogo{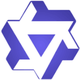}{Qwen3-VL-Plus} & 8.6 & 25.6 & 22.5 & 37.8 & 14.3 & 39.1 & 27.5 & 51.1 & 0.0 & 10.0 & 16.5 & 35.4 \\
\modellogo{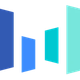}{Seed-1.8} & 11.4 & 33.5 & 22.5 & 34.6 & 22.9 & 48.7 & 27.5 & 51.2 & 10.0 & 18.8 & 20.0 & 39.0 \\
\modellogo{qwen}{Qwen3-VL-235B-A22B} & 5.7 & 23.2 & 17.5 & 29.5 & 8.6 & 33.3 & 27.5 & 50.4 & 0.0 & 3.6 & 13.5 & 30.8 \\
\modellogo{qwen}{Qwen3-VL-30B-A3B} & 2.9 & 20.3 & 20.0 & 36.5 & 2.9 & 26.1 & 27.5 & 51.1 & 0.0 & 3.5 & 12.4 & 30.6 \\
\modellogo{qwen}{Qwen3.5-2B} & 0.0 & 5.6 & 5.8 & 10.6 & 2.9 & 9.4 & 7.5 & 28.4 & 0.0 & 9.1 & 3.7 & 13.3 \\
\modellogo{qwen}{Qwen3.5-4B} & 4.8 & 9.4 & 2.5 & 6.3 & 4.8 & 21.6 & 1.7 & 8.2 & 5.0 & 12.7 & 3.5 & 11.3 \\
\modellogo{qwen}{Qwen3.5-9B} & 10.5 & 28.8 & 5.0 & 17.6 & 12.4 & 35.3 & 17.5 & 44.1 & 0.0 & 9.5 & 10.0 & 28.8 \\

\midrule
\multicolumn{13}{l}{\cellcolor{bggray}\textcolor{textgray}{\textbf{\textit{Ours}}}}\\
\modellogo{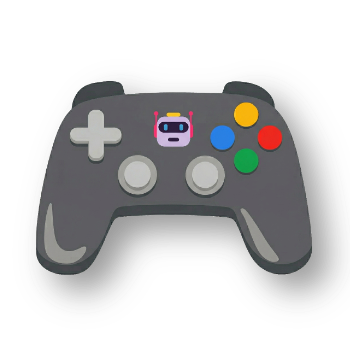}{\textsc{LightAct}-2B} & 7.6 & 28.7 & 17.5 & 30.3 & 18.1 & 49.5 & 15.8 & 46.1 & 6.7 & 24.0 & 13.9 & 36.9 \\
\modellogo{lightact}{\textsc{LightAct}-4B} & 17.1 & 36.4 & 15.0 & 31.0 & 23.8 & 54.5 & 20.8 & 49.7 & 6.7 & 23.7 & 17.6 & 40.5 \\
\modellogo{lightact}{\textsc{LightAct}-9B} & 14.3 & 34.3 & 21.7 & 36.6 & 26.7 & 59.2 & 19.2 & 48.7 & 3.3 & 20.9 & 18.4 & 41.8 \\

\bottomrule
\end{tabular}
}
}

\label{tab:main}
\end{table*}

\endgroup

\subsection{RL-Based Policy Internalization}
\label{sec:policytraining}
After collection, we pair each decision context in $\D_{\mathrm{root}}$ with its constructed reward function $r_i:\A_e\to[0,1]$:
\begin{equation}
 \D_{\mathrm{reward}}=
 \{(x_i,r_i)\mid(s_i,x_i)\in\D_{\mathrm{root}}\}.
\end{equation}
This converts the action comparisons in source experience into policy-learning examples: $x_i$ supplies the task and visual history, while $r_i$ provides feedback on the selected action. Starting from $\theta=\theta_0$, we train $\pi_\theta$ to select valuable actions directly from $x_i$.

The policy $\pi_\theta$ generates a response $y$ from the decision context $x_i$, which is parsed into an action $a=\mathcal P(y)$ when valid. A response is valid when it satisfies the public action syntax and argument constraints and selects an action in $\A_e$. The response reward is
\begin{equation}
 R_i(y)=
 \begin{cases}
 r_i(a), & \text{if }y\text{ is valid},\\
 0, & \text{otherwise}.
 \end{cases}
 \label{eq:training_reward}
\end{equation}
Sampling examples uniformly from $\D_{\mathrm{reward}}$, we optimize the expected response reward with reference-policy regularization:
\begin{equation}
\begin{aligned}
 \mathcal J_{\mathrm{RL}}(\theta)
 &=\E_{(x_i,r_i)\sim\D_{\mathrm{reward}}}\Big[\\
 &\quad\E_{y\sim \pi_\theta(\cdot\mid x_i)}[R_i(y)]\\
 &\quad-\lambda D_{\mathrm{KL}}\!\left(
 \pi_\theta(\cdot\mid x_i)\,\Vert\,\pi_0(\cdot\mid x_i)\right)\Big],
\end{aligned}
\label{eq:rl_objective}
\end{equation}
where $\lambda\geq0$ controls deviation from the initial policy $\pi_0$. The current policy samples responses conditioned on $x_i$ and receives $R_i(y)$; the collected contexts and reward maps remain fixed during these updates. Established policy-gradient algorithms optimize this objective, with their configurations specified in the experiments.

After learning, the policy generates $y_t\sim\pi_{\theta^\star}(\cdot\mid x_t)$ and executes the parsed action $a_t=\mathcal P(y_t)$ using the task description, visual observations, and action interface of the environment being played.

\begin{figure*}[t]%
\centering%
\includegraphics[width=\textwidth]{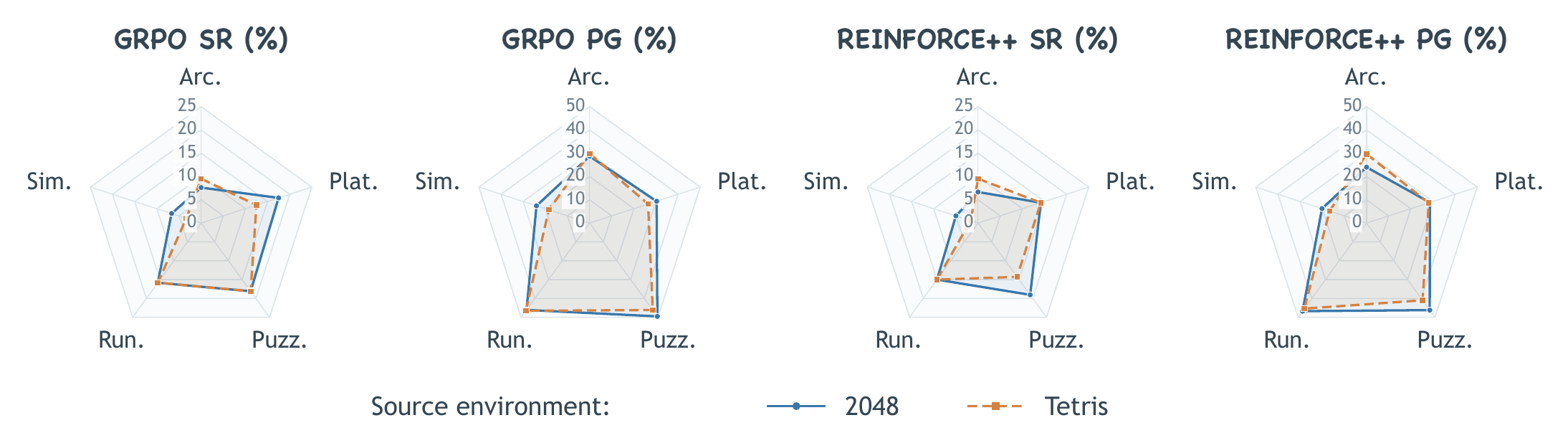}%
\caption{\textbf{Source-environment ablation.} GameWorld SR and PG for 2B policies trained on 2048 or Tetris with GRPO and REINFORCE++.}%
\label{fig:source_environment}%
\end{figure*}%

\section{Experiments}
\label{sec:experiments}

\subsection{Benchmark and Evaluation Setting}
We evaluate \textsc{LightAct} on GameWorld~\cite{gameworld}, which comprises 34 games and 170 tasks across five categories: Arcade, Platformer, Puzzle, Runner, and Simulation. Each run has a budget of 100 decisions, with the environment paused during inference. At each decision, the policy receives the game instructions, the current screenshot, and two rounds of image--text interaction history, and selects an action from the game's interface. Action duration, episode termination, and resets follow the protocol. Published GameWorld results use greedy decoding with one run per task. Our Qwen3.5 and \textsc{LightAct} evaluations use temperature 1.0, top-$p$ 0.95, a 2,048-token output cap, and three runs per task. We use GameWorld's state-based evaluator to measure success rate (SR), the fraction of successful runs, and progress (PG), normalized completion toward task-specific targets.

\subsection{Implementation Details}
We instantiate \textsc{LightAct} with Qwen3.5-2B, Qwen3.5-4B, and Qwen3.5-9B~\cite{qwen35}. At each scale, experience collection, branch continuation, and skill extraction use frozen copies of the same model that initializes the trainable policy. Acting uses the native non-thinking mode. For each source game--task pair, we collect trajectories within a budget of 10,000 decisions and retain up to 1,024 deduplicated roots, with seed 42 for the initial reset. We equip source games with snapshot support through \texttt{exportSnapshot()} and \texttt{importSnapshot()}, restoring each decision root before evaluating candidate actions. Collection and branch continuation use temperature 1.0, top-$p$ 0.95, and a 2,048-token output cap. Branch valuation uses $H=10$ including the initial action and $K=4$ continuations, processing eight roots concurrently. Rewards use environment-reported score gains with per-root min--max normalization. We train with GRPO~\cite{grpo}, REINFORCE++~\cite{reinforcepp}, or DAPO~\cite{dapo} in verl/HybridFlow~\cite{verl}. Tables~\ref{tab:configuration} and~\ref{tab:training_configuration} in the appendix give the collection and reward settings and the training hyperparameters, respectively.

\subsection{Baselines and Main Comparisons}
\paragraph{\textsc{LightAct} achieves competitive progress against proprietary models.}
\textsc{LightAct} brings compact policies close to strong proprietary models in benchmark progress (Table~\ref{tab:main}). At 4B, it reaches 40.5\% overall PG, comparable to GPT-5.2 at 40.6\%; at 9B, it reaches 41.8\%, approaching Gemini-3-Flash-Preview at 41.9\%. On Puzzle tasks, \textsc{LightAct}-9B achieves 59.2\% PG, exceeding the reported progress of both models. The remaining gap is clearer in task completion, where overall SR reaches 18.4\% for \textsc{LightAct}-9B compared with 21.2\% for Gemini. By comparing actions from a shared state, \textsc{LightAct} turns interaction outcomes into supervision for policy learning. These results suggest that internalizing such experience enables compact VLMs to achieve competitive game-playing performance through direct policy inference.

\paragraph{\textsc{LightAct} consistently improves policy performance across model scales.}
\textsc{LightAct} increases both SR and PG in all five game categories at 2B, 4B, and 9B, indicating benefits for task completion as well as partial progress. The overall PG gains are 23.6, 29.2, and 13.0 percentage points, respectively. The 4B policy gains most on Runner tasks, where PG rises from 8.2\% to 49.7\%, even though its training experience comes from a puzzle game. Notably, \textsc{LightAct}-2B surpasses the pretrained 9B baseline in both PG (36.9\% versus 28.8\%) and SR (13.9\% versus 10.0\%), showing that adaptation through experience can improve the effectiveness of a smaller policy. The 9B model also gains 8.4 percentage points in SR over its initialization, demonstrating that stronger pretrained performance still leaves room for improvement through supervision from interaction.

\begingroup
\definecolor{bggray}{HTML}{F2F2F2}
\definecolor{textgray}{HTML}{4A4A4A}
\begin{table}[t]
\small
\centering
\caption{\textbf{Policy internalization.} GameWorld PG (\%) for base and trained policies across three model scales and RL algorithms.}
\label{tab:algorithm}
\setlength{\tabcolsep}{2pt}
\renewcommand{\arraystretch}{1.02}
\setlength{\cmidrulekern}{2pt}

\begin{tabularx}{\linewidth}{@{}l*{5}{>{\centering\arraybackslash}X}|>{\centering\arraybackslash}X@{}}
\toprule
\multirow{2}{*}{\textbf{Model}} &
\textbf{Arc.} & \textbf{Plat.} & \textbf{Puzz.} &
\textbf{Run.} & \textbf{Sim.} & \textbf{Overall} \\
\cmidrule(lr){2-2}\cmidrule(lr){3-3}\cmidrule(lr){4-4}\cmidrule(lr){5-5}\cmidrule(lr){6-6}\cmidrule(lr){7-7}
 & $\mathcal{PG}$ & $\mathcal{PG}$ & $\mathcal{PG}$ & $\mathcal{PG}$ & $\mathcal{PG}$ & $\mathcal{PG}$ \\
\midrule
\multicolumn{7}{l}{\cellcolor{bggray}\textcolor{textgray}{\textbf{\textit{Qwen3.5-2B}}}}\\
Base & 5.6 & 10.6 & 9.4 & 28.4 & 9.1 & 13.3 \\
+REINFORCE++ & 24.0 & 28.6 & 46.2 & 46.8 & 20.1 & 34.5 \\
+GRPO & 28.7 & 30.3 & 49.5 & 46.1 & 24.0 & 36.9 \\
+DAPO & 26.6 & 29.9 & 50.2 & 45.0 & 20.6 & 35.9 \\
\midrule
\multicolumn{7}{l}{\cellcolor{bggray}\textcolor{textgray}{\textbf{\textit{Qwen3.5-4B}}}}\\
Base & 9.4 & 6.3 & 21.6 & 8.2 & 12.7 & 11.3 \\
+REINFORCE++ & 29.7 & 30.9 & 51.9 & 48.9 & 19.5 & 37.9 \\
+GRPO & 36.4 & 31.0 & 54.5 & 49.7 & 23.7 & 40.5 \\
+DAPO & 33.8 & 30.0 & 56.3 & 49.1 & 26.3 & 40.2 \\
\midrule
\multicolumn{7}{l}{\cellcolor{bggray}\textcolor{textgray}{\textbf{\textit{Qwen3.5-9B}}}}\\
Base & 28.8 & 17.6 & 35.3 & 44.1 & 9.5 & 28.8 \\
+REINFORCE++ & 36.3 & 35.5 & 53.1 & 49.0 & 11.2 & 39.6 \\
+GRPO & 34.3 & 36.6 & 59.2 & 48.7 & 20.9 & 41.8 \\
+DAPO & 33.1 & 32.4 & 59.7 & 49.7 & 19.4 & 40.7 \\
\bottomrule
\end{tabularx}
\end{table}
\endgroup

\subsection{Ablations and Diagnostics}
\paragraph{Policy internalization yields consistent gains across RL algorithms.}
We compare GRPO, REINFORCE++, and DAPO, using the same reward tables for each model scale (Table~\ref{tab:algorithm}). All nine trained configurations improve overall PG over their base policies, with gains ranging from 10.8 to 29.2 percentage points. This pattern suggests that the action-level rewards provide useful supervision across different policy-update rules. The optimizer still affects how these gains are distributed across games: at 4B, DAPO achieves higher Puzzle and Simulation PG than GRPO, while GRPO attains higher overall PG. Thus, the common benefit of learning from action outcomes persists alongside algorithm-dependent differences in performance across categories.

\begin{figure}[t]
\centering
\includegraphics[width=\linewidth]{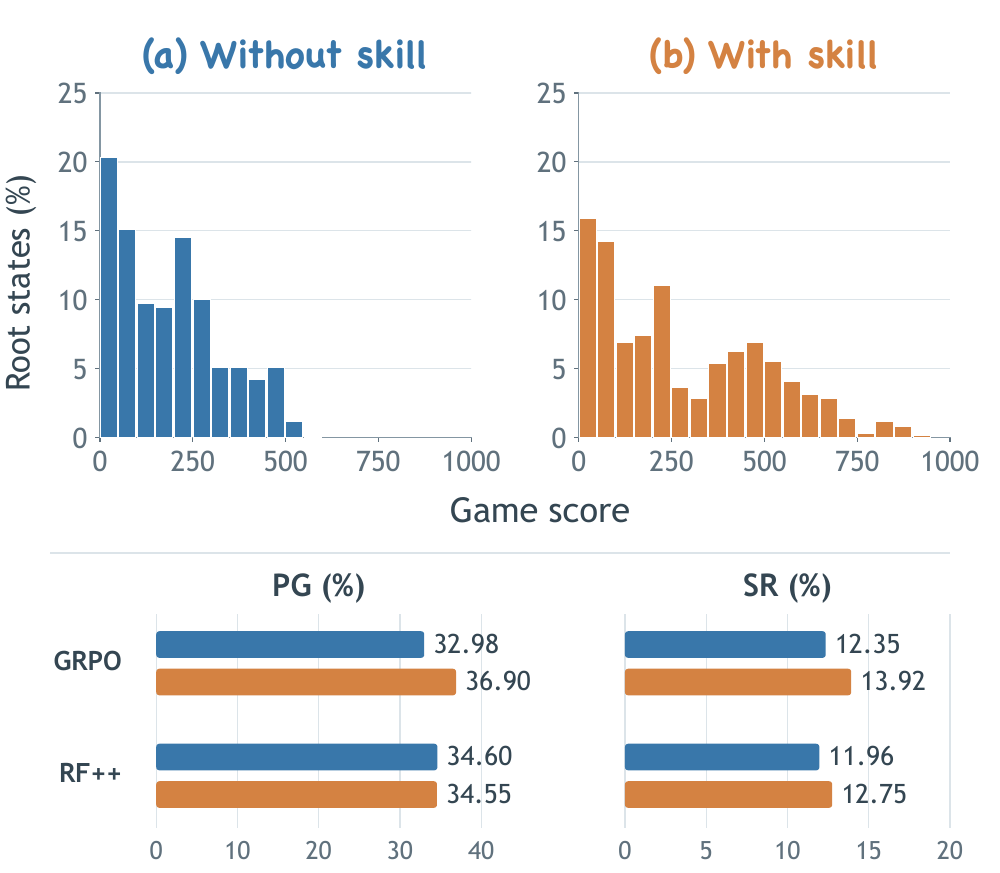}
\caption{\textbf{Skill-guided exploration.} Top: score distributions of 1,024 roots in 2048 (50-point bins). Bottom: GameWorld PG and SR for 2B policies trained with GRPO or REINFORCE++ (RF++).}
\label{fig:skill_root_distribution}
\end{figure}

\begin{figure*}[!t]
\centering
\includegraphics[width=\textwidth]{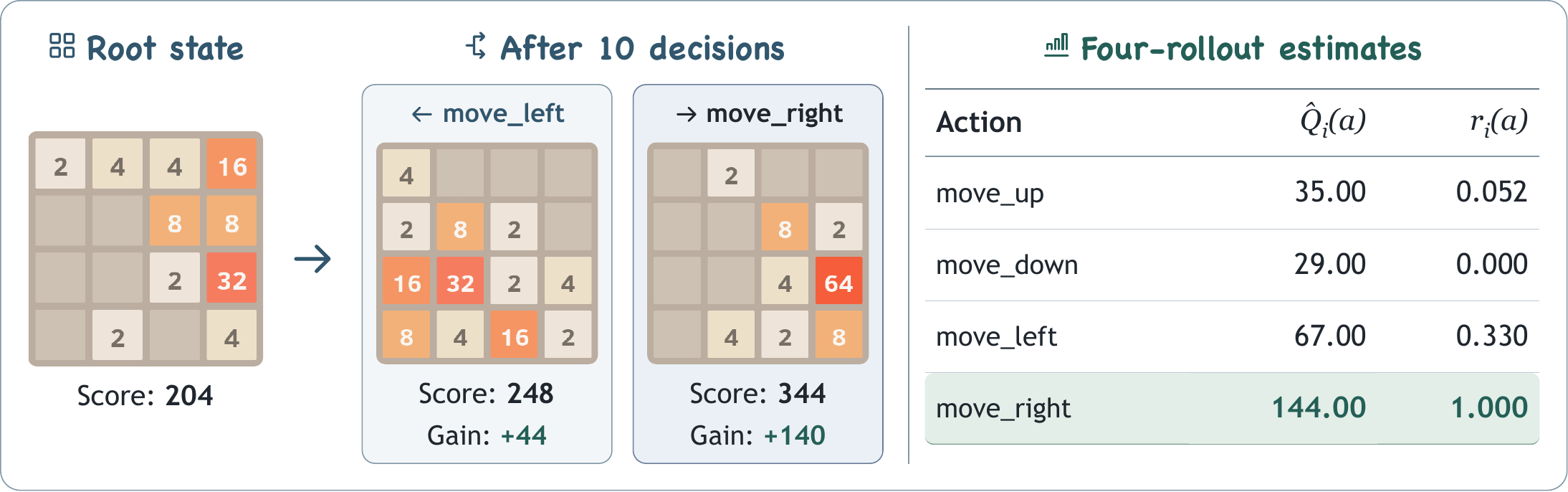}
\caption{\textbf{From a root state to action rewards in 2048.} Left and right gain 24 immediately but yield different outcomes. The table averages four ten-decision continuations per action and normalizes gains into rewards; endpoints are nearest each action's mean gain.}
\label{fig:root_reward_2048}
\end{figure*}

\begin{figure}[!t]
\centering
\includegraphics[width=\columnwidth]{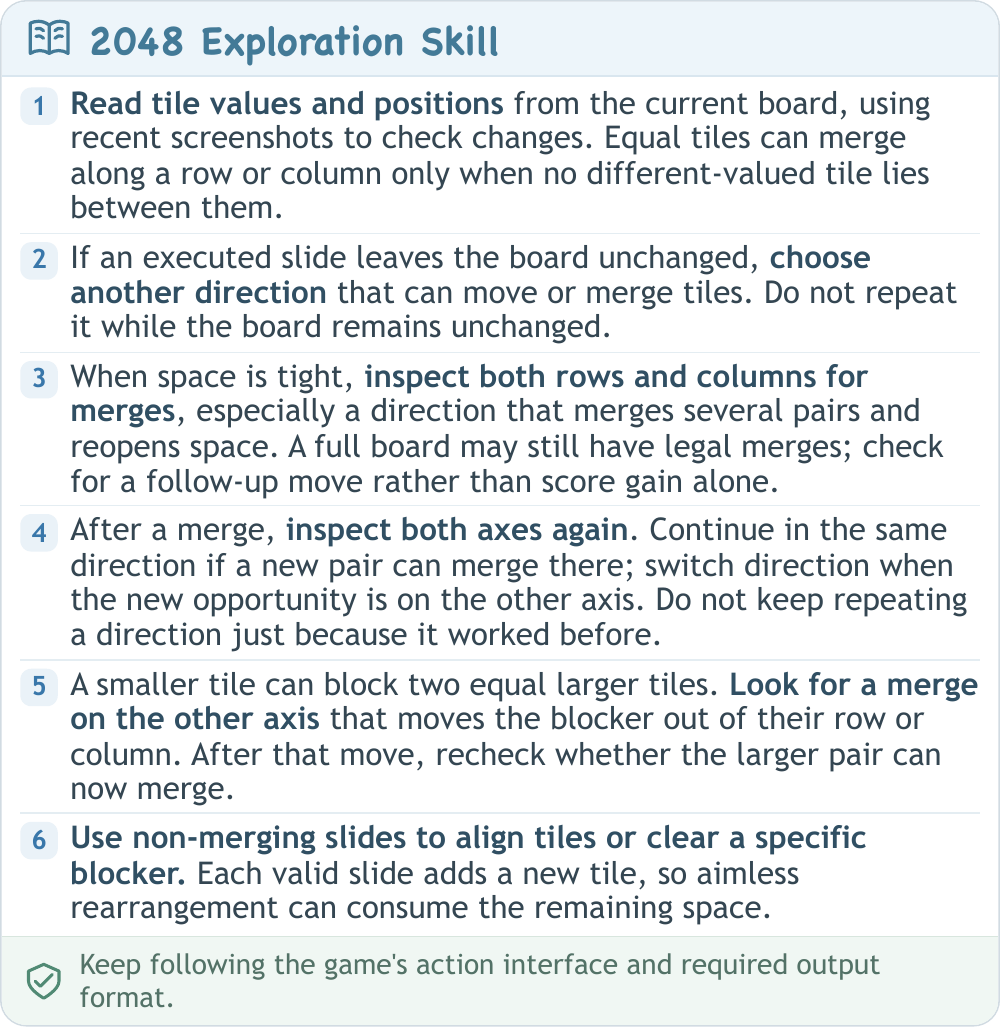}
\caption{\textbf{2048 exploration skill.} The textual guidance used during root-state collection, with action-selection principles highlighted.}
\label{fig:2048_skill}
\vspace{-8pt}
\end{figure}

\paragraph{Skill guidance broadens the score distribution of collected roots.}
Figure~\ref{fig:skill_root_distribution} compares 1,024 roots collected with and without skill guidance in 2048; Figure~\ref{fig:2048_skill} shows the skill used during collection. By drawing on interaction experience, the skill guides the policy away from repeated ineffective moves and helps it explore a broader range of game states. Guided collection extends the score range from 0--552 to 0--944 and raises the mean score from 186.6 to 280.2. The proportion of roots scoring at least 500 increases from 1.3\% to 19.4\%, while lower-score states remain represented. With GRPO, skill guidance raises PG from 32.98\% to 36.90\% and SR from 12.35\% to 13.92\% (lower panels of Figure~\ref{fig:skill_root_distribution}). With REINFORCE++, PG changes little (34.60\% to 34.55\%), while SR rises from 11.96\% to 12.75\%. These results suggest that skill guidance broadens the experience available for policy learning, with benefits that depend on the optimizer.

\paragraph{Source-game experience supports generalization across game categories.}
With Tetris as the source game, both GRPO and REINFORCE++ improve PG over the 2B base policy across all five categories (Figure~\ref{fig:source_environment}). Arcade PG, for example, rises from 5.6\% to 29.8\% and 29.6\%, respectively. These gains extend beyond the source game's Puzzle category, suggesting that policy internalization helps \textsc{LightAct} apply experience to different game mechanics. Experience from a single game can provide supervision for broader visual decisions. The varying gains across categories indicate that transfer is uneven, leaving room to improve the breadth of behaviors learned from source experience.

\paragraph{Rollout outcomes distinguish actions with identical immediate gains.}
In Figure~\ref{fig:root_reward_2048}, moving left or right from the same root yields an immediate score gain of 24, but the resulting boards support different continuations. The rightward branch produces a 64 tile, whereas the leftward branch retains a maximum tile of 32. Across four rollouts, moving right achieves a mean gain of 144 versus 67 for moving left, yielding normalized rewards of 1.000 and 0.330. The rollout comparison turns delayed action outcomes into a training preference that immediate score gains alone cannot provide.

\section{Conclusion}
We presented \textsc{LightAct}, a framework that turns interaction experience into guidance for exploration and supervision for policy learning. Game-specific skills broaden root-state coverage, while short rollouts from shared roots reveal the relative utility of candidate actions. Experiments on GameWorld show consistent improvements across model scales and policy optimizers, with compact policies achieving competitive progress against proprietary models. These findings demonstrate the value of organizing experience around states and action outcomes for learning game-playing policies.

\section{Limitations}
Generalization across source environments remains limited, and the current framework does not directly support action spaces with continuous parameters. Our study focuses on short-horizon decisions; long-horizon decision-making remains to be validated. Future work will investigate integration with multi-turn reinforcement learning and evaluation on games that require sustained planning. Reward shaping currently relies on scores read from environment states. We will further explore purely visual self-improvement that derives learning signals from observations without access to internal environment states.

\section*{LLM Usage Statement}
OpenAI Codex was used to assist with code generation and completion, and OpenAI GPT models assisted with related-literature discovery. Codex also assisted with drafting and editing manuscript text. The authors led the research process, problem formulation, computational experiments, and factual verification, and take full responsibility for the scientific claims, results, references, and final manuscript.

{\small\bibliographystyle{ieeenat_fullname}\bibliography{references}}
\onecolumn
\appendix
\setcounter{figure}{0}
\setcounter{table}{0}
\renewcommand{\thefigure}{S\arabic{figure}}
\renewcommand{\thetable}{S\arabic{table}}
\renewcommand{\theHfigure}{appendix.\arabic{figure}}
\renewcommand{\theHtable}{appendix.\arabic{table}}
\begin{figure*}[p]
\centering
\pdfbookmark[1]{Appendix}{appendix-start}
{\Large\bfseries Appendix\par}\medskip
\begingroup
\setlength{\fboxsep}{7pt}
\setlength{\fboxrule}{0.5pt}
\newsavebox{\caseleft}
\newsavebox{\caseright}
\newlength{\casewidth}
\newlength{\caseheight}
\setlength{\casewidth}{\dimexpr0.49\textwidth-2\fboxsep-2\fboxrule\relax}
\newcommand{\casefont}{\fontfamily{ppl}\fontsize{9}{10.2}\selectfont\raggedright}
\sbox{\caseleft}{\begin{minipage}[t]{\casewidth}\casefont
\textbf{System input}\par\smallskip
You are an expert game agent specialized in playing video games. Your goal is to play the game and achieve the task goal.\par
Observe the current game screen to identify your character and key objects. Your action must follow the game rules and the instructions of the current task. Execute the actions frame-by-frame.\par
You do NOT have direct access to keyboard or mouse actions. You must act through the registered semantic control list only.\par
\smallskip
\textbf{\# Game Rules}\par
You are playing 2048, a sliding tile puzzle game.\par
\smallskip
\textbf{\#\# Game Objective}\par
- Combine matching tiles to create higher values.\par
\smallskip
\textbf{\#\# Game Rules}\par
- All tiles slide in the direction you press.\par
- When two tiles with the same number collide, they merge into one tile with doubled value.\par
- After each valid move (any tile is moved), a new tile of 2 or 4 appears in a random empty cell.\par
- Game ends when no more moves are possible (board is full and no merges available).\par
\smallskip
\smallskip
\textbf{\# Role and Controls}\par
You control the 2048 board.\par
Choose exactly one action per step to slide tiles.\par
\smallskip
REGISTERED ACTIONS (Semantic Controls).\par
Choose exactly ONE action per step:\par
\smallskip
- `move\_up`: Slide all tiles up.\par
- `move\_down`: Slide all tiles down.\par
- `move\_left`: Slide all tiles left.\par
- `move\_right`: Slide all tiles right.\par
\smallskip
\textbf{\# Task Instruction}\par
Combine tiles with minimal board clutter and reach value 64 or higher.\par
\smallskip
\smallskip
\textbf{\# Output Format}\par
Action: one short sentence describing what to do.\par
Return exactly one function call per step in this format with no suffix:\par
\texttt{\detokenize{<tool_call>}}\par
\texttt{\detokenize{<function=<function-name>>}}\par
\texttt{\detokenize{<parameter=<argument-name>>}}\par
argument-value\par
\texttt{\detokenize{</parameter>}}\par
\texttt{\detokenize{</function>}}\par
\texttt{\detokenize{</tool_call>}}\par
\smallskip
Use a registered semantic action as the function name. Repeat the <parameter> block for each required argument, or omit it when the action has no arguments.\par
\end{minipage}}
\sbox{\caseright}{\begin{minipage}[t]{\casewidth}\casefont
\textbf{Historical messages}\par\smallskip
\begin{minipage}[t]{0.48\linewidth}\vspace{0pt}
\textbf{User (step 44)}\par Game screen:\par
\includegraphics[width=\linewidth,trim=440bp 205bp 440bp 115bp,clip]{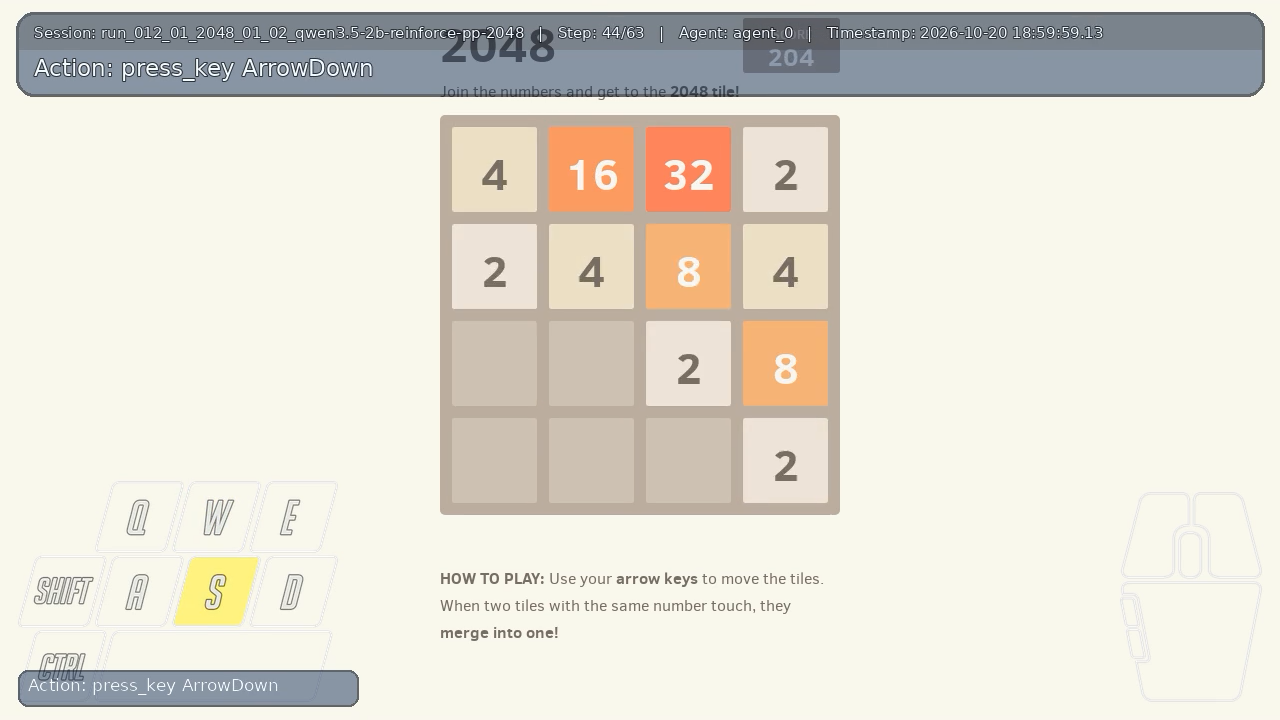}\par\smallskip
\textbf{Assistant}\par\smallskip
To create more opportunities for merges, the board can be moved down again.\par\smallskip
\texttt{\detokenize{<tool_call>}}
\texttt{\detokenize{<function=move_down>}}
\texttt{\detokenize{</function>}}
\texttt{\detokenize{</tool_call>}}\par
\end{minipage}\hfill
\begin{minipage}[t]{0.48\linewidth}\vspace{0pt}
\textbf{User (step 45)}\par Game screen:\par
\includegraphics[width=\linewidth,trim=440bp 205bp 440bp 115bp,clip]{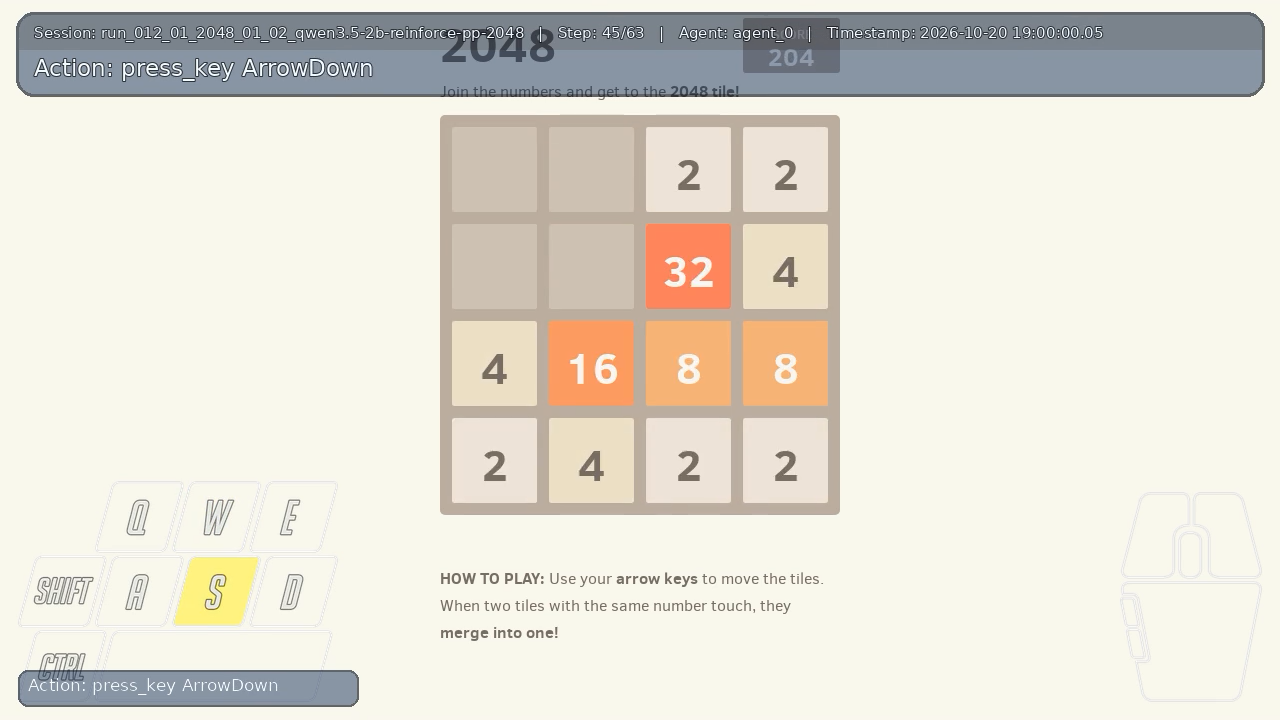}\par\smallskip
\textbf{Assistant}\par\smallskip
To create more opportunities for merges, the board can be moved down again.\par\smallskip
\texttt{\detokenize{<tool_call>}}
\texttt{\detokenize{<function=move_down>}}
\texttt{\detokenize{</function>}}
\texttt{\detokenize{</tool_call>}}\par
\end{minipage}\par\medskip\hrule\medskip
\textbf{Current user message (step 46)}\par Game screen:\par\smallskip
{\centering\includegraphics[width=0.42\linewidth,trim=440bp 205bp 440bp 115bp,clip]{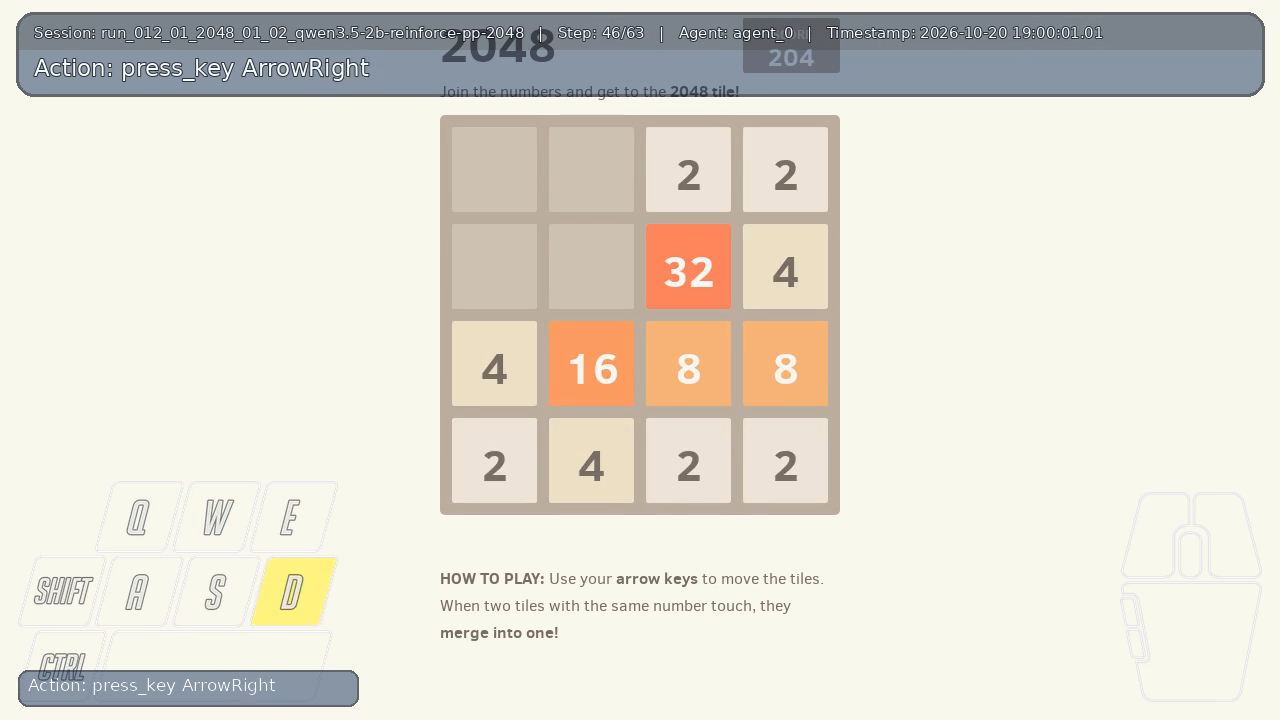}\par}
\textbf{Model output}\par\smallskip
The game screen doesn't change after repeated move\_down actions, so it is best to try a different direction to create new opportunities for merges.\par\smallskip
\texttt{\detokenize{<tool_call>}}
\texttt{\detokenize{<function=move_right>}}
\texttt{\detokenize{</function>}}
\texttt{\detokenize{</tool_call>}}\par
\medskip\hrule\medskip\textbf{Observed outcome}\par\smallskip
\begin{minipage}[c]{0.36\linewidth}\includegraphics[width=\linewidth,trim=440bp 205bp 440bp 115bp,clip]{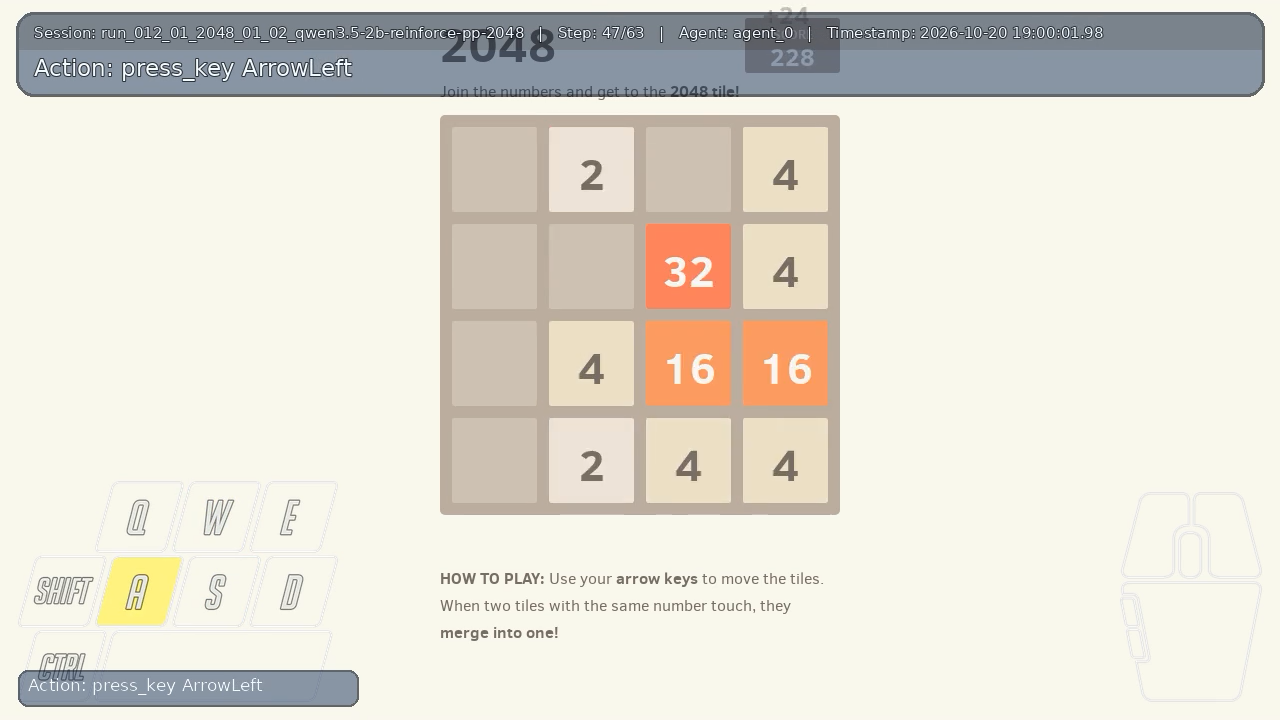}\end{minipage}\hfill
\begin{minipage}[c]{0.60\linewidth}\raggedright
\texttt{move\_right} merges $2+2$, $8+8$, and $2+2$.\par\smallskip Score: $204\rightarrow228$ ($+24$).
\end{minipage}
\end{minipage}}
\setlength{\caseheight}{\dimexpr\ht\caseleft+\dp\caseleft\relax}
\ifdim\dimexpr\ht\caseright+\dp\caseright\relax>\caseheight
\setlength{\caseheight}{\dimexpr\ht\caseright+\dp\caseright\relax}
\fi
\newcommand{\caseframe}[1]{\fcolorbox{black!25}{black!3}{\begin{minipage}[t][\caseheight][t]{\casewidth}\usebox{#1}\end{minipage}}}
\caseframe{\caseleft}\hfill\caseframe{\caseright}
\endgroup
\caption{\textbf{A 2048 decision by \textsc{LightAct}-2B trained with REINFORCE++} (task 01\_02, repetition 3, step 46). After a repeated downward move leaves the board unchanged, the policy selects \texttt{move\_right}, increasing the score from 204 to 228.}
\label{fig:2048_case}
\end{figure*}

\clearpage
\begin{figure*}[p]
\centering
\begingroup
\setlength{\fboxsep}{5pt}
\setlength{\fboxrule}{0.5pt}
\newcommand{\mariofont}{\fontfamily{ppl}\fontsize{9}{10}\selectfont\raggedright}
\newcommand{\mariocall}[1]{\texttt{\detokenize{<tool_call> <function=}#1\detokenize{> </function> </tool_call>}}}
\newcommand{\mariopanel}[1]{\fcolorbox{black!25}{black!3}{%
\begin{minipage}{\dimexpr\textwidth-2\fboxsep-2\fboxrule\relax}\mariofont #1\end{minipage}}}
\mariopanel{%
\textbf{System input (game-specific excerpt)}\par\smallskip
\begin{minipage}[t]{0.485\linewidth}\mariofont\vspace{0pt}
\textbf{\# Game Rules}\par
You are playing Super Mario Bros.\par
Control Mario through platforming levels, collecting coins and defeating enemies.\par\smallskip
\textbf{\#\# Game Objective}\par
- Reach the flagpole at the end of each level.\par
- Collect coins from question blocks or kill enemies for points.\par\smallskip
\textbf{\#\# Game Rules}\par
- You have 3 lives and limited time per level.\par
- Touching enemies from the side or below kills Mario. Jump on enemies to kill enemies.\par
- Jumping underneath question blocks will reveal coins or power-ups.\par
- Power-ups: Mushroom make Mario grow big.
\end{minipage}\hfill
\begin{minipage}[t]{0.485\linewidth}\mariofont\vspace{0pt}
\textbf{\# Role and Controls}\par
You control Mario.\par
Navigate through platforming levels by running, jumping, and avoiding hazards.\par\smallskip
REGISTERED ACTIONS (Semantic Controls).\par
Choose exactly ONE action per step:\par
- `wait`: Stand still and wait.\par
- `move\_right`: Walk right.\par
- `move\_left`: Walk left.\par
- `jump`: Jump straight up.\par
- `jump\_right`: Jump while moving right.\par
- `jump\_left`: Jump while moving left.\par
- `duck`: Duck/crouch (when big).\par\smallskip
\textbf{\# Task Instruction}\par
Reach at least 400 score in one run.
\end{minipage}}
\par\smallskip
\mariopanel{%
\textbf{Historical messages}\par\smallskip
\begin{minipage}[t]{0.485\linewidth}\mariofont\vspace{0pt}
\textbf{User (step 38)}\quad Game screen:\par\smallskip
\includegraphics[width=\linewidth,trim=0bp 260bp 0bp 96bp,clip]{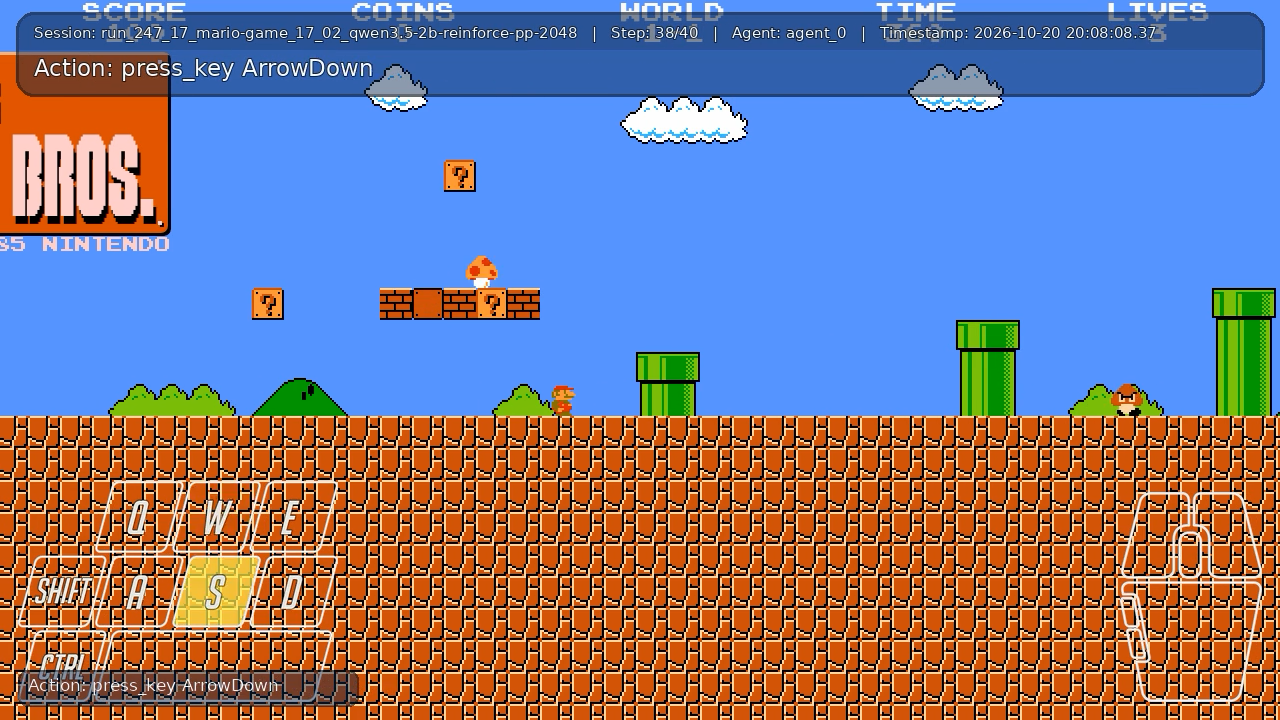}\par\smallskip
\textbf{Assistant}\par\smallskip
I am jumping on the block to move forward, but the screen shows the jump didn't register. I need to be more precise. The Goomba is moving towards me, so I should duck to get below it and skip it.\par\smallskip
\mariocall{duck}
\end{minipage}\hfill
\begin{minipage}[t]{0.485\linewidth}\mariofont\vspace{0pt}
\textbf{User (step 39)}\quad Game screen:\par\smallskip
\includegraphics[width=\linewidth,trim=0bp 260bp 0bp 96bp,clip]{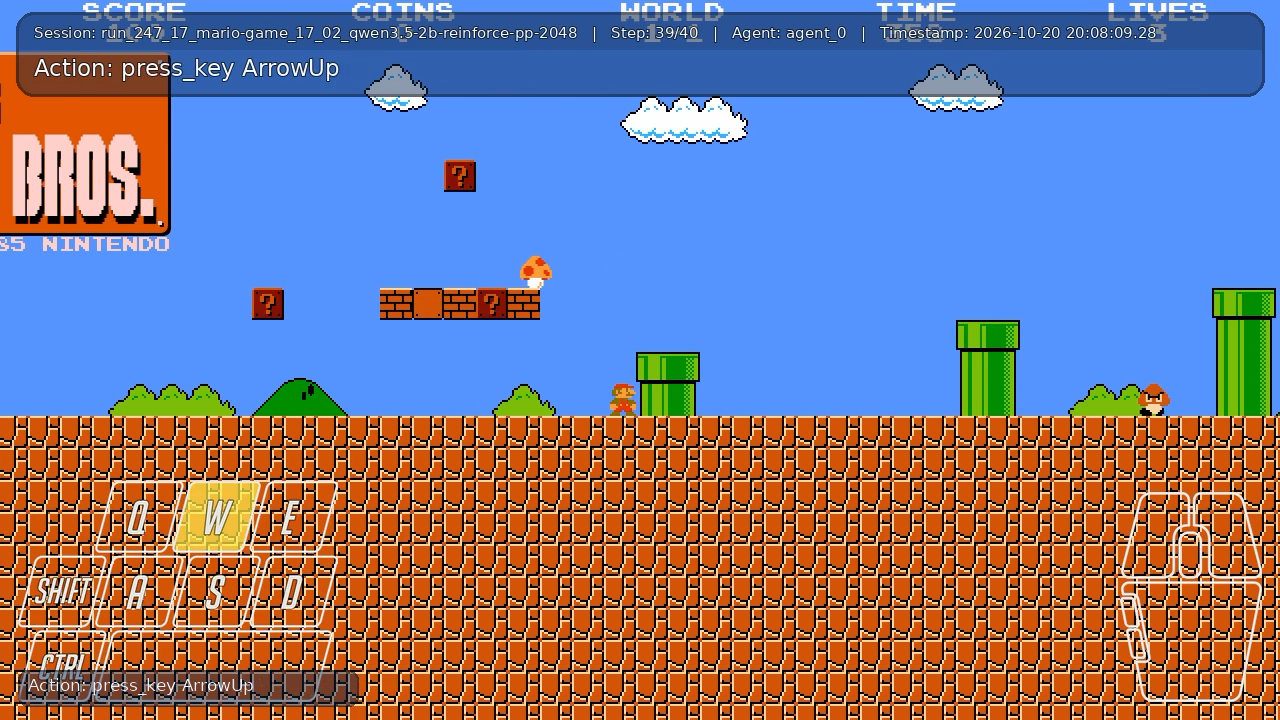}\par\smallskip
\textbf{Assistant}\par\smallskip
You are looking for a coin to get points. There's a question block that should contain a coin. Jump on it to collect the coin.\par\smallskip
\mariocall{jump}
\end{minipage}}
\par\smallskip
\mariopanel{%
\textbf{Current user message (step 40)}\quad Game screen:\par\smallskip
{\centering\includegraphics[width=0.76\linewidth,trim=0bp 260bp 0bp 96bp,clip]{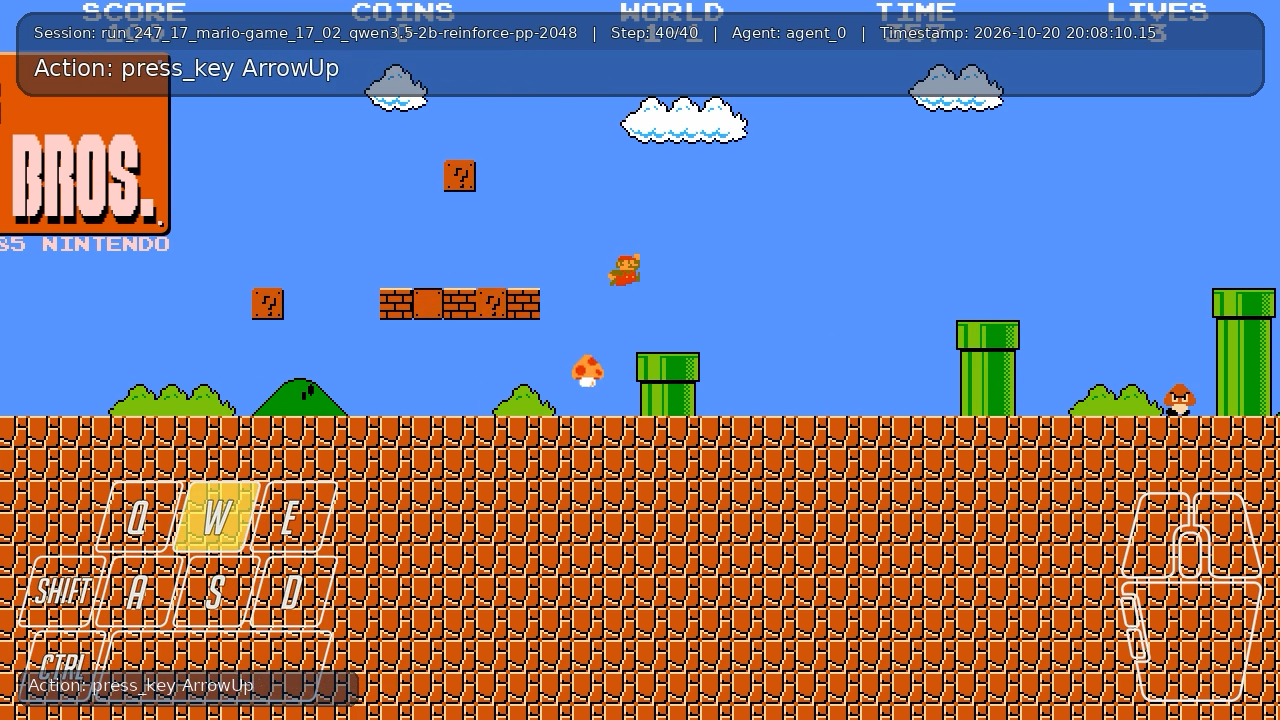}\par}
\smallskip\textbf{Model output}\par\smallskip
You have a mushroom, which makes you grow big, to help pass gaps. The mushroom is near a question block.\par\smallskip
\mariocall{jump}\par
\medskip\hrule\medskip
\textbf{Observed outcome}\quad Score: $100\rightarrow1{,}100$ ($+1{,}000$). Mario collects the mushroom and grows from small to big, completing the 400-point task.
}
\endgroup
\caption{\textbf{A Mario decision by \textsc{LightAct}-2B trained with REINFORCE++ on 2048} (task 17\_02, repetition 2, step 40). The recorded \texttt{jump} action is followed by mushroom collection and task completion. Shared system instructions appear in Fig.~\ref{fig:2048_case}.}
\label{fig:mario_case}
\end{figure*}

\clearpage
\begin{figure*}[p]
\centering
\begingroup
\setlength{\fboxsep}{7pt}
\setlength{\fboxrule}{0.5pt}
\newcommand{\rootcasefont}{\fontfamily{ppl}\fontsize{10}{12}\selectfont}
\newcommand{\rootcasepanel}[1]{\fcolorbox{black!25}{black!3}{%
\begin{minipage}{\dimexpr\textwidth-2\fboxsep-2\fboxrule\relax}\rootcasefont #1\end{minipage}}}
\newcommand{\boardt}[1]{\includegraphics[height=155pt,trim=519bp 215bp 513bp 10bp,clip]{figures/case_root_reward/#1}}
\rootcasefont
\textbf{One root}\quad$\longrightarrow$\quad\textbf{Different initial actions}\quad$\longrightarrow$\quad\textbf{Rollout outcomes and rewards}\par\bigskip
\rootcasepanel{%
\textbf{Tetris}\hfill $K=4$, $H=10$\par\medskip
\begin{minipage}[c]{0.18\linewidth}\centering
\textbf{Root state}\par\smallskip
\boardt{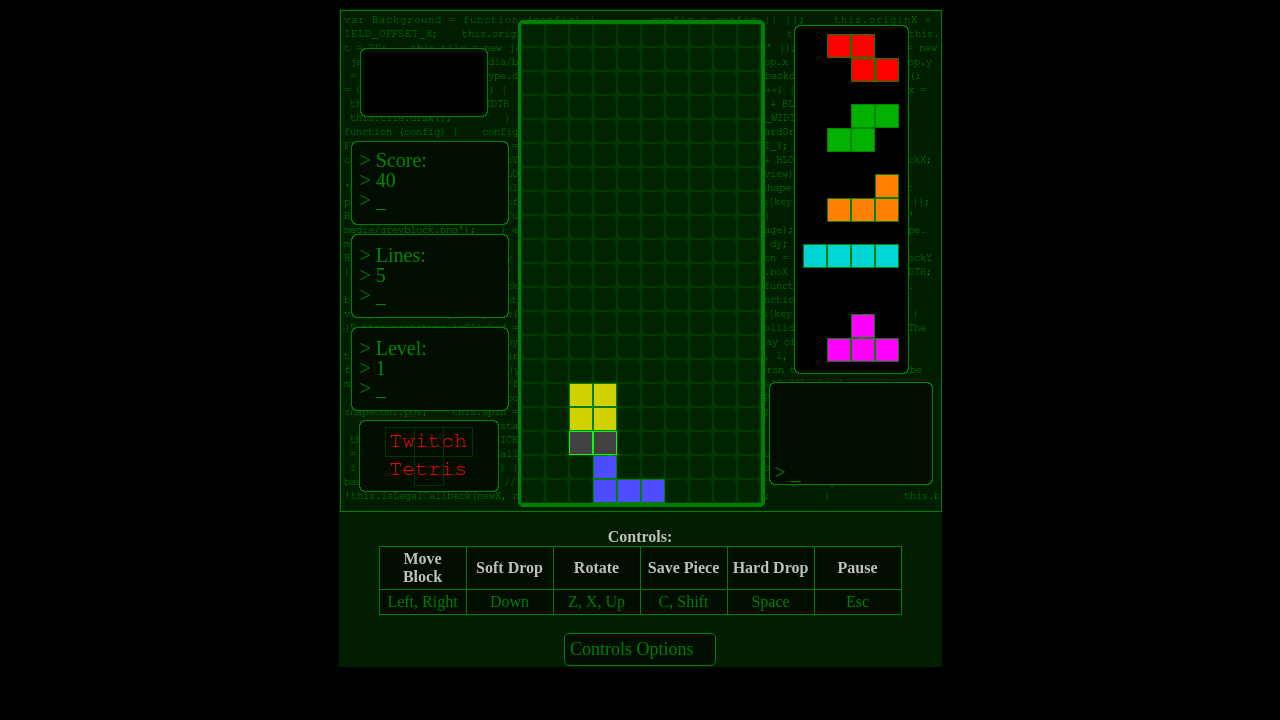}\par\smallskip
Score: $49$
\end{minipage}\hfill
\begin{minipage}[c]{0.04\linewidth}\centering$\longrightarrow$\end{minipage}\hfill
\begin{minipage}[c]{0.39\linewidth}\centering
\textbf{After 10 decisions}\par\medskip
\begin{minipage}[t]{0.48\linewidth}\centering
\texttt{move\_left}\par\smallskip
\boardt{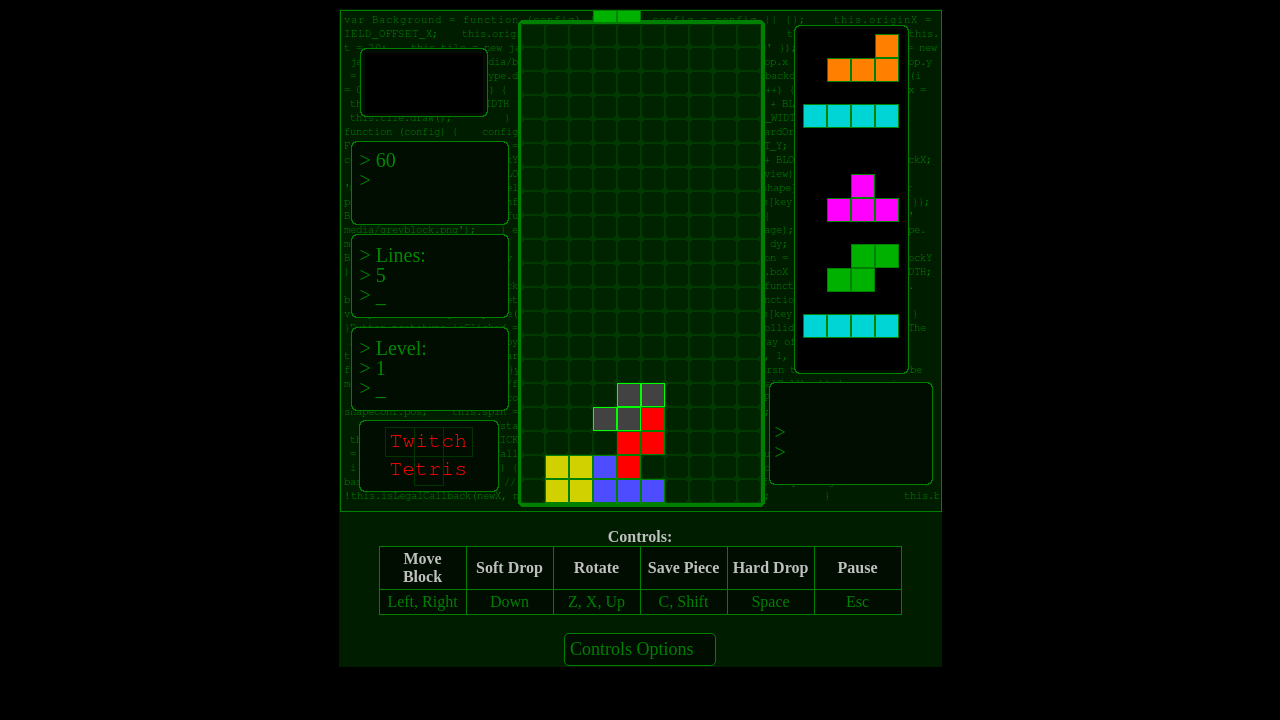}\par\smallskip
Score: $75$\par
Gain: $+26$
\end{minipage}\hfill
\begin{minipage}[t]{0.48\linewidth}\centering
\texttt{rotate\_ccw}\par\smallskip
\boardt{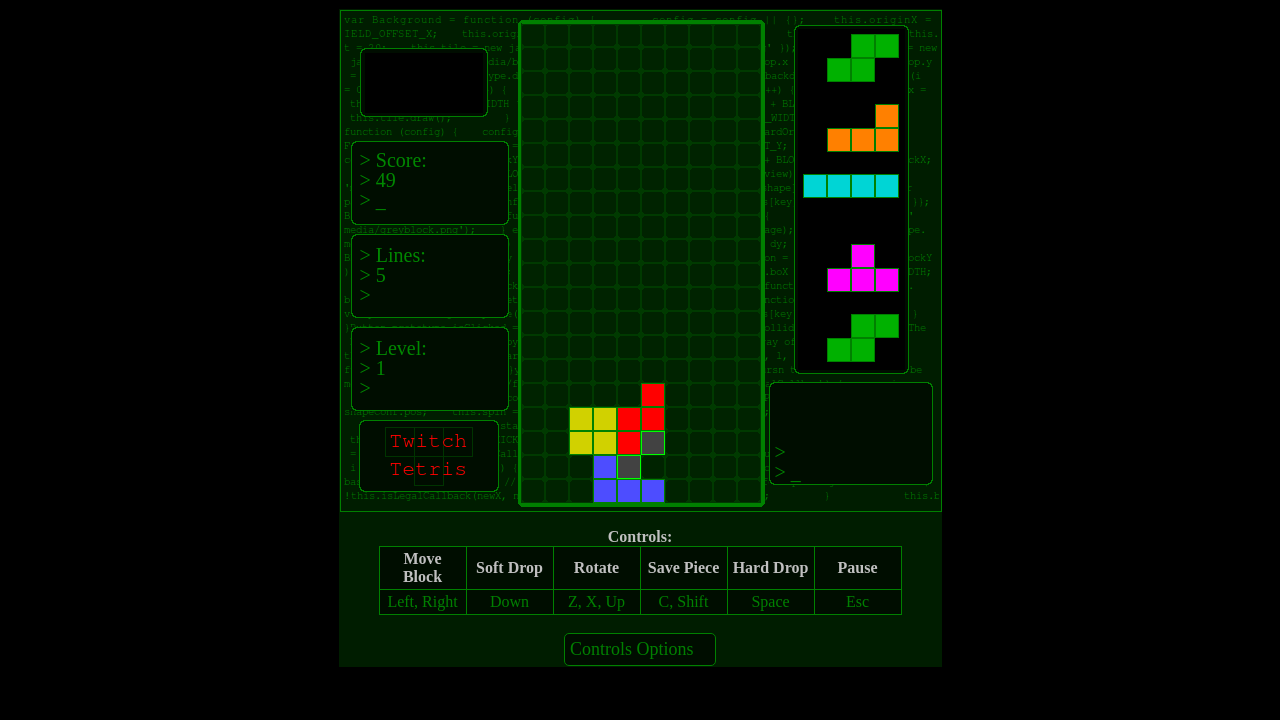}\par\smallskip
Score: $58$\par
Gain: $+9$
\end{minipage}
\end{minipage}\hfill
\begin{minipage}[c]{0.34\linewidth}\centering
\textbf{Four-rollout estimates}\par\medskip
\setlength{\tabcolsep}{3pt}
\renewcommand{\arraystretch}{1.22}
\begin{tabular}{@{}lrr@{}}
\toprule
\textbf{Action} & $\widehat Q_i(a)$ & $r_i(a)$\\
\midrule
\texttt{wait} & 18.25 & 0.357\\
\texttt{move\_left} & 24.00 & 0.631\\
\texttt{move\_right} & 19.50 & 0.417\\
\texttt{soft\_drop} & 17.25 & 0.310\\
\texttt{rotate\_cw} & 16.25 & 0.262\\
\texttt{rotate\_ccw} & 10.75 & 0.000\\
\texttt{hard\_drop} & 23.00 & 0.583\\
\rowcolor{black!8}\texttt{hold\_piece} & \textbf{31.75} & \textbf{1.000}\\
\bottomrule
\end{tabular}
\end{minipage}\par\medskip
The leftward branch places the O piece on the floor beside the stack. In the rotation branch, it settles above the existing blocks, leaving empty cells beneath it.
}
\par\bigskip
\rootcasefont
\textbf{Action value:} mean score gain across four rollouts.\quad
\textbf{Reward:} min--max normalized action value.
\endgroup
\caption{\textbf{From a root state to action rewards in Tetris.} Two initial actions are illustrated with rollout endpoints selected closest to their respective mean gains. Scores and gains beneath the images describe the displayed rollouts; the table aggregates four ten-decision continuations per action under the frozen policy.}
\label{fig:root_reward_case}
\end{figure*}

\clearpage

\begingroup
\definecolor{bggray}{HTML}{F2F2F2}
\definecolor{textgray}{HTML}{4A4A4A}
\begin{table*}[t]
\centering
\small
\renewcommand{\arraystretch}{1.12}
\caption{\textbf{Trajectory collection and reward construction.}}
\label{tab:configuration}
\setlength{\tabcolsep}{0.25mm}

\begin{tabularx}{\textwidth}{@{}p{0.38\textwidth}X@{}}
\toprule
\textbf{Parameter} & \textbf{Setting} \\
\midrule
\multicolumn{2}{l}{\cellcolor{bggray}\textcolor{textgray}{\textbf{\textit{Source collection}}}} \\
Runs / decision budget & 1 per game--task pair / 10,000 steps \\
Episode decision limit & 100 \\
Root-state limit & 1,024 per run \\
Deduplication & Environment state \\
Visual context & Current image + 2 image--text history rounds \\
Skill guidance & Fixed game-specific skill \\
Skill extractor & Frozen initial model $\pi_0$ \\
Temperature / top-$p$ & 1.0 / 0.95 \\
Max. response tokens & 2,048 \\
\midrule
\multicolumn{2}{l}{\cellcolor{bggray}\textcolor{textgray}{\textbf{\textit{Branch exploration}}}} \\
Candidate actions & All public actions \\
Repetitions per action $K$ & 4 \\
Branch horizon $H$ & 10 decisions (including initial action) \\
Continuation policy & Frozen initial model $\pi_0$ \\
Concurrent root states & 8 \\
\midrule
\multicolumn{2}{l}{\cellcolor{bggray}\textcolor{textgray}{\textbf{\textit{Reward construction}}}} \\
Score source & Environment state \\
Action value & Mean score gain over $K$ continuations \\
Action-value normalization & Per-state min--max \\
\bottomrule
\end{tabularx}
\end{table*}

\endgroup

\paragraph{Collection details.}
At each model scale, $\pi_0$ denotes the corresponding frozen Qwen3.5 initialization. In 2048, deduplication uses the ordered grid of tile values, including empty cells, and retains the first occurrence across episodes. Collection continues after task success and resets at termination or after 100 decisions. During collection and branch continuation, an invalid response consumes one decision step without executing an action or advancing game time.

\begingroup
\definecolor{bggray}{HTML}{F2F2F2}
\definecolor{textgray}{HTML}{4A4A4A}
\begin{table*}[t]
\centering
\small
\renewcommand{\arraystretch}{1.12}
\caption{\textbf{Policy-training configuration.} Settings for GRPO, REINFORCE++, and DAPO in verl.}
\label{tab:training_configuration}
\setlength{\tabcolsep}{0.25mm}

\begin{tabularx}{\textwidth}{@{}p{0.32\textwidth}*{3}{>{\raggedright\arraybackslash}X}@{}}
\toprule
\textbf{Parameter} & \textbf{GRPO} & \textbf{REINFORCE++} & \textbf{DAPO} \\
\midrule
\multicolumn{4}{l}{\cellcolor{bggray}\textcolor{textgray}{\textbf{\textit{Training settings}}}} \\
Base model & Qwen3.5-2B/4B/9B & Qwen3.5-2B/4B/9B & Qwen3.5-2B/4B/9B \\
Freeze vision encoder & Yes & Yes & Yes \\
Batch size (prompts) & 64 & 64 & 64 \\
Minibatch size (prompts) & 64 & 64 & 64 \\
Learning rate & $5\times10^{-7}$ & $5\times10^{-7}$ & $5\times10^{-7}$ \\
Training epochs & 2 & 2 & 2 \\
Actor update epochs & 1 & 1 & 1 \\
Max. prompt / response tokens & 8,192 / 1,024 & 8,192 / 1,024 & 8,192 / 1,024 \\
Temperature / top-$p$ & 1.0 / 1.0 & 1.0 / 1.0 & 1.0 / 1.0 \\
Strategy / precision & FSDP / BF16 & FSDP / BF16 & FSDP / BF16 \\
\midrule
\multicolumn{4}{l}{\cellcolor{bggray}\textcolor{textgray}{\textbf{\textit{Algorithm settings}}}} \\
Responses per prompt & 5 & 1 & 5 \\
Advantage normalization & Group & Batch & Group \\
KL regularization & Actor loss & Reward penalty & None \\
KL coefficient & 0.01 & 0.001 (fixed) & 0 \\
Loss reduction & Sequence mean of token means & Token mean & Token mean \\
Discount factor $\gamma$ & -- & 1.0 & -- \\
Clip ratio (lower / upper) & 0.2 / 0.2 & 0.2 / 0.2 & 0.2 / 0.28 \\
Entropy coefficient & 0 & 0 & 0 \\
Reward-group filtering & -- & -- & Enabled \\
Overlong penalty & -- & -- & Disabled \\
\bottomrule
\end{tabularx}
\end{table*}

\endgroup

\end{document}